%% file: main.tex
\documentclass{article}
\usepackage{style/iclr2027_conference,times}
\usepackage[T1]{fontenc}
\usepackage[utf8]{inputenc}
\usepackage{amsmath,amssymb,booktabs,multirow,graphicx,microtype}
\usepackage{tikz}
\usepackage{colortbl}
\definecolor{oursrow}{HTML}{F4EBF4}
\usepackage{placeins}
\usepackage{wrapfig}
\usepackage{needspace}
\usepackage{caption}
\usepackage{adjustbox}
\usetikzlibrary{arrows.meta,positioning,calc,fit}
\usepackage{hyperref}
\usepackage{url}
\hypersetup{colorlinks=true,linkcolor=blue,citecolor=blue,urlcolor=blue,pdftitle={SenseAlign: Understanding Vision and Touch through Language at Test Time}}
\graphicspath{{figures/}}
\title{SenseAlign: Understanding Vision and Touch through Language at Test Time}
\input{authors}
\newcommand{\method}{SenseAlign}
\iclrfinalcopy
\begin{document}
\raggedbottom
\maketitle
\lhead{Preprint}
\begin{abstract}
Tactile--visual--language models combine vision and touch through a shared
language space to recognize objects and materials. In real-world deployment,
changes in lighting, contact quality or sensor hardware can shift visual or
tactile inputs and degrade recognition performance. When both modalities
contribute to a prediction, an unreliable modality can override correct
evidence from the other. Since labeled data are generally unavailable
during deployment, test-time adaptation offers a way to recover
recognition performance from unlabeled observations.
In this setting, we find that comparing accumulated observations with the
text embeddings of candidate classes reveals modality reliability differences
that confidence can miss. Refining features against competing classes also
improves recognition beyond geometric alignment alone.
These findings motivate SenseAlign, which uses a common semantic reference
for test-time modality weighting and feature correction. Alignment Trust
weights each modality according to its support for shared predicted classes.
Anchor Realignment uses the same class text embeddings to fit a shared
feature correction that separates predicted classes from alternatives.
Both modules use the same unlabeled history while keeping
all encoders frozen. Extensive experiments across diverse datasets
demonstrate strong recognition performance under both natural shifts
and controlled corruptions.
\end{abstract}

\input{sections/01_introduction}
\input{sections/02_motivation}

\input{sections/03_problem_setting}
\input{sections/03_method}
\input{sections/04_related_work}
\input{sections/05_experiments}
\input{sections/06_conclusion}
\label{page:main_end}
\clearpage
\flushbottom

\section*{Reproducibility statement}
The appendix specifies the class vocabulary and prompts, calibration solver,
frozen encoder state, online update order, and benchmark protocols.
It identifies the sampling units and aggregation rules for each comparison.
The accompanying code regenerates the figures and the appendix tables from
saved evaluation outputs and checks
every cell of the main tables against those outputs.

\section*{Ethics statement}
The study evaluates recognition on existing visual and tactile datasets. It does
not collect new human data or deploy a physical robot. Recognition gains under
image corruptions do not establish safe manipulation: contact force, action
selection, and physical failure recovery require separate evaluation. Dataset
licenses and redistribution constraints apply to any accompanying data release.

\section*{Statement on the use of large language models}
Large language model tools assisted with drafting, source retrieval, code
inspection, experiment coordination and figure preparation.
Quantitative figures use recorded evaluations; the method diagram
illustrates the stated procedure. The authors are responsible for the
scientific and editorial content.

\bibliography{references,related_additions}
\bibliographystyle{style/iclr2027_conference}
\clearpage
\appendix
\input{appendix/protocol_details}
\FloatBarrier
\input{appendix/tact_method_details}
\FloatBarrier
\input{appendix/tact_evaluation}
\FloatBarrier
\input{appendix/unitouch_baselines}
\FloatBarrier
\input{appendix/unitouch_tagc}
\FloatBarrier
\FloatBarrier
\input{appendix/matched_readout_controls}
\FloatBarrier

\input{appendix/tact_secondary}
\FloatBarrier
\input{appendix/realignment_diagnostic}
\FloatBarrier
\input{appendix/discussion}
\end{document}

%% file: authors.tex
\author{Chuyang Ye$^{1}$ \quad Qinting Jiang$^{2}$ \quad Jingyan Jiang$^{3}$\thanks{Corresponding author.}\\[3pt]
\normalfont\small $^{1}$New York University\\
\normalfont\small $^{2}$Tsinghua Shenzhen International Graduate School, Tsinghua University\\
\normalfont\small $^{3}$Shenzhen Technology University}
\hypersetup{pdfauthor={Chuyang Ye; Qinting Jiang; Jingyan Jiang}}

%% file: sections/01_introduction.tex
\section{Introduction}
\label{sec:introduction}

Tactile--visual--language (TVL) models align visual and tactile observations
with language, bringing what a robot sees and feels into a shared semantic
space~\citep{fu2024tvl,yang2024unitouch,feng2025anytouch}.
For a robot sorting materials, this connects two different views of the
same object. Touch can distinguish surfaces that look alike, while vision
reveals information beyond a small contact patch. Textual descriptions
provide a common vocabulary for these observations, allowing both senses
to contribute to recognizing objects and their physical properties.

These two views are acquired under different sensing conditions.
Poor lighting can obscure visual distinctions, while incomplete contact
or a different sensor can weaken tactile information. During recognition,
each modality scores candidate classes by comparing its representation
with their text descriptions; fusion combines these scores into a prediction.
Our analysis finds cases where equal fusion loses distinctions that one
modality still preserves, with either vision or touch taking the stronger
role. Other cases benefit from both senses. The challenge is therefore
to retain their complementarity as their relative value changes.

Responding to such changes without labels is the premise of test-time
adaptation (TTA). Existing approaches learn from prediction entropy and
cross-modal information~\citep{wang2021tent,niu2023sar,yang2024read,guo2025sumi},
including methods that address reliability bias and modality-specific
shift~\citep{yang2024read,chen2025tsa}.
Prediction confidence offers an intuitive signal for deciding which sense
to trust. It measures how strongly a model favors its prediction over the
alternatives, but a strong preference can persist when the prediction is
wrong. Our analysis finds precisely this mismatch: the less accurate
modality can be more confident. Giving it greater influence can reinforce
its errors. For fusion, the relevant question is whether each modality
still preserves the class distinctions needed for recognition under the
current sensing conditions.

The shared language space provides a reference for examining these class
distinctions. Its text embeddings, or semantic anchors, describe the same
candidate classes even when the visual and tactile observations change.
We group accumulated paired observations under shared predicted classes,
then compare each modality's support for those classes against the
alternatives. Using the same observations and assignments lets us examine
how the two modalities represent the same evidence. Our analysis finds
that this semantic comparison favors the more accurate modality even
when confidence reverses their ordering. Comparing support against the same anchors offers a basis for
adjusting their contributions without assuming that one sense should
always dominate.

Adjusting these contributions changes how branch predictions are combined,
but leaves each branch's feature--anchor relationships unchanged.
Those relationships determine how a branch distinguishes candidate
classes, providing a complementary target for adaptation. The same semantic
anchors can therefore guide both modality weighting and feature correction.
Building on these two roles, we introduce \method\ for test-time adaptation.
\emph{Alignment Trust} adjusts modality weights through semantic
support, while \emph{Anchor Realignment} fits a shared feature correction
for class discrimination. Both use the same unlabeled history with frozen
encoders.

Our contributions are as follows.
\begin{enumerate}
\item We identify confidence reversals in TVL recognition under changing
  sensing conditions. Shared text embeddings offer an alternative comparison
  signal when the less accurate modality remains more confident.
\item We introduce \method, coupling modality weighting and feature
  realignment through a common semantic reference. Both use one unlabeled
  history to adapt modality contributions and class discrimination with
  frozen encoders.
\item Experiments across two TVL models cover natural shifts and controlled
  corruptions. Matched comparisons show average gains beyond confidence-based
  fusion, while component studies demonstrate the benefit of combining
  weighting and feature correction.
\end{enumerate}

\FloatBarrier

%% file: sections/02_motivation.tex
\Needspace{24\baselineskip}
\section{Motivation}
\label{sec:motivation}

\input{figures/scene_teaser}
\input{figures/regimes}

\paragraph{Complementarity depends on the sensing condition.}
Neither modality is uniformly more informative. When one loses a semantic
distinction that the other retains, equal fusion can move observations
across the wrong class boundary. The upper and middle examples in
Figure~\ref{fig:regimes}a show this interference in opposite directions:
vision separates classes that touch confuses in one case, while touch
separates classes that vision confuses in the other.

Both modalities can also preserve useful distinctions, as shown by
the stream-level comparison in Figure~\ref{fig:regimes}b. A useful adaptation rule must
therefore adjust their contributions without assuming that one source
should always dominate. This requires a signal for comparing the two
modalities without knowing which one is more accurate.

\paragraph{Semantic support provides a reference beyond confidence.}
Confidence alone can favor the less accurate modality. It measures how
strongly a prediction favors one class, which need not reflect whether
that class is correct. The lower example in Figure~\ref{fig:evidence}b
shows this mismatch: the more confident branch has substantially lower
accuracy. Increasing its contribution would strengthen the wrong signal.

Text embeddings provide a common semantic reference for comparing
the observations. We group the two modalities under the same estimated
labels, then compare their accumulated representations with the assigned
anchors and the alternatives. Using identical groups keeps differences
in sample selection from entering the modality comparison.

Relative semantic support can reveal distinctions that confidence misses.
In the upper and lower examples of Figure~\ref{fig:evidence}, alignment
favors the more accurate modality, including the case where confidence
reverses the ordering. Randomizing the memory
labels weakens this structure, linking the signal to the feature--label
correspondence.

\paragraph{The same reference can guide feature correction.}
Weighting changes the contribution of each branch, but leaves its feature
positions relative to the text anchors unchanged. A branch given more
weight still makes the same distinctions among candidate classes.
The shared groups also let us examine these distinctions and use their
assigned anchors as targets for feature correction. Such a correction
must consider competing anchors, because recognition depends on relative
class scores. This motivates two uses of the same semantic reference:
comparing modality contributions and improving the class distinctions in
the features they supply.

\input{figures/evidence}

%% file: figures/scene_teaser.tex
\begin{wrapfigure}[22]{r}{0.40\textwidth}
  \centering
  \input{figures/scene_teaser_artwork}
  \caption{\textbf{Vision and touch under sensing and hardware shifts.}
  Natural changes in lighting or contact can affect the modalities unequally.
  The upper panel uses brightness and defocus corruptions to illustrate
  three possible modality relationships. The lower panel shows separate
  wire contacts recorded by GelSight Mini, DIGIT and DuraGel, illustrating
  differences across tactile sensor hardware and capture conditions.}
  \label{fig:scene-teaser}
  
\end{wrapfigure}
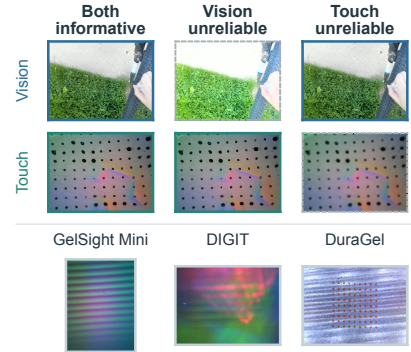

%% file: figures/scene_teaser_artwork.tex
\includegraphics[width=\linewidth]{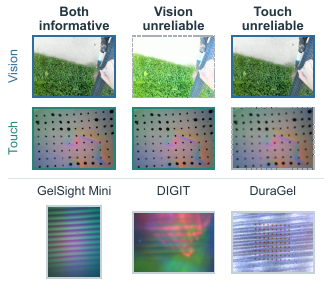}

%% file: figures/regimes.tex
\begin{figure}[t]
\centering
\includegraphics[width=.95\linewidth]{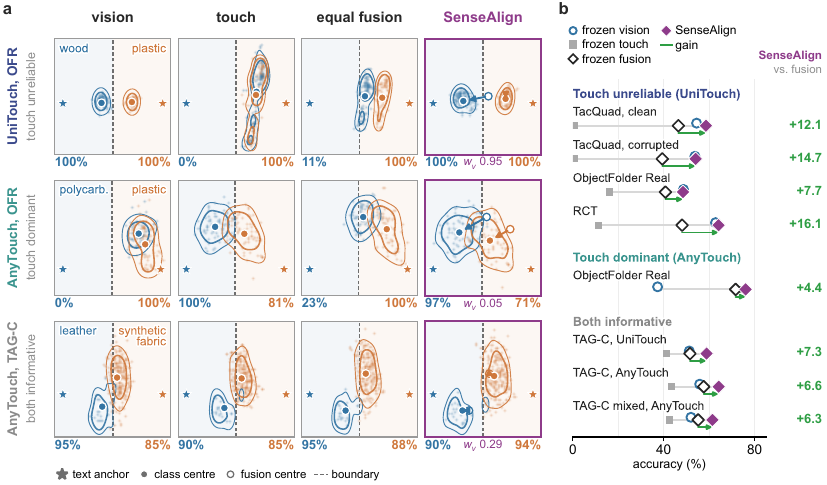}
\caption{\textbf{Modality reliability varies with backbone and stream.}
(a) Selected class pairs projected along their text-embedding decision direction.
Stars mark text anchors; percentages give each class's share on its own
side of the dashed boundary, not full-vocabulary accuracy.
Arrows show shifts from equal-fusion to \method\ class centres;
$w_v$ is the mean vision weight.
(b) Stream-level accuracy; green arrows and labels give gains over frozen
equal fusion in percentage points.}
\label{fig:regimes}
\end{figure}

%% file: figures/evidence.tex
\begin{figure}[t]
\centering
\includegraphics[width=.90\linewidth]{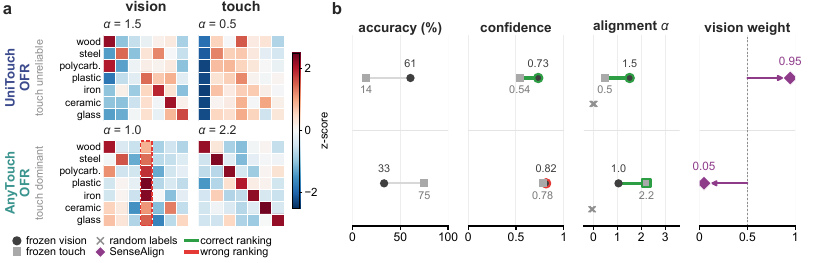}
\caption{\textbf{Semantic alignment distinguishes modalities when confidence misranks them.}
(a) Standardized scores between pseudo-label group means and text embeddings. Columns follow the row-label order;
grey rows lack accepted observations. The red box highlights the visual
branch's strong scores for plastic across groups.
(b) Frozen accuracy, mean top-1 probability, final-memory alignment
$\alpha$ and mean online vision weight.
Green and red outlines indicate whether confidence or alignment favors
the more or less accurate modality; crosses show alignment with randomized
memory labels. Purple diamonds show \method's vision weights.}
\label{fig:evidence}
\end{figure}

%% file: sections/03_problem_setting.tex
\section{Problem setting}
\label{sec:problem-setting}

Multimodal test-time adaptation updates a pretrained model from unlabeled
target observations under distribution shift.
Let $\mathcal D_0$ denote the source distribution and $\mathcal D_b$ the test distribution
at batch $b$. A shift may affect vision alone:
$\mathcal D_b(x^v)\neq \mathcal D_0(x^v)$ while
$\mathcal D_b(x^t)=\mathcal D_0(x^t)$. It may instead affect touch alone, or both
modalities. The modalities may shift at different times. Text descriptions
specify the vocabulary $\mathcal C$ of $C$ classes. The frozen text encoder maps these descriptions to semantic anchors.

We observe ordered batches $\mathcal B_b=\{(x_i^v,x_i^t)\}_{i\in I_b}$,
with both inputs describing the same recognition target. Adaptation aims
to improve online classification using only
the pretrained model, class descriptions and the unlabeled stream.
True labels, the affected modality, shift identities and change boundaries
are unavailable to adaptation; $\widehat y_i$ denotes a pseudo-label. We predict each batch before updating. Model and adaptation state persist
within streams and reset between streams.

%% file: sections/03_method.tex
\section{SenseAlign}
\label{sec:method}

\method\ uses one semantic reference to compare modality contributions
and correct the features being combined. Alignment Trust gives more weight
to the modality whose accumulated observations better support their assigned
text anchors. Anchor Realignment uses those anchors as targets for a
shared feature correction. Both operations draw on the same groups of
past observations, connecting fusion and correction through one shared
memory (Figure~\ref{fig:method-architecture}). All encoders remain frozen.

\input{figures/method_architecture}

\subsection{A shared semantic reference}
\label{sec:semantic-reference}

Comparing modalities requires a common basis for judging their semantic
support. We therefore group the same observations under shared pseudo-labels
and sample weights. Their branch means support the modality comparison;
their fused means supply the fitting signal for feature correction.
For modality $m\in\{v,t\}$, let
$u_i^m=\nu(h_0^m(x_i^m))$ be the frozen unit feature of observation $i$,
where $\nu(z)=z/\max(\|z\|_2,\epsilon)$ and features are row vectors.
Encoders, prompts and classifier remain unchanged during adaptation.

Each class $c$ has one text prompt. Its normalized embedding $a_c$
serves both as the classifier anchor and as the target for feature
correction. The rows of $A\in\mathbb R^{C\times d}$ are these anchors.

To keep the groups independent of adapted fusion weights, we assign
labels from the raw equal-fusion feature $u_i^0=\nu((u_i^v+u_i^t)/2)$.
Let $c(z)=\arg\max_j za_j^\top$ and let $\delta(z)$ be the gap between
its largest and second-largest cosine scores. An observation enters
memory only when its top score and margin are positive:
\begin{equation}
 \begin{aligned}
 \widehat y_i&=c(u_i^0),\qquad
 \gamma_i=\mathbf 1\!\left[u_i^0 a_{\widehat y_i}^\top>0,\ \delta(u_i^0)>0\right],\\
 \lambda_i&=\gamma_i\delta(u_i^0).
 \end{aligned}
 \label{eq:memory_selection}
\end{equation}
Larger margins give more weight to assignments that separate the leading
class from its closest competitor; rejected samples have zero weight.
These labels remain model-generated assignments, rather than ground-truth
supervision. Both modalities use the same assignments and weights.

Before batch $b$, let $\mathcal H_{c,b}$ contain earlier observations
assigned pseudo-label $c$. Each modality and their equal-fusion feature
use the same samples and weights to form a weighted mean:
\begin{equation}
 \mu_{c,b}^m=\frac{\sum_{j\in\mathcal H_{c,b}}\lambda_j u_j^m}{\Lambda_{c,b}},
 \qquad \Lambda_{c,b}=\sum_{j\in\mathcal H_{c,b}}\lambda_j,
 \quad m\in\{v,t,0\}.
 \label{eq:class_means}
\end{equation}
Only groups with $\Lambda_{c,b}>0$ enter the set $\mathcal C_b$.
The memory stores weighted sums and total weights, rather than past
samples. Its raw features are never replaced by corrected features
(see the appendix).

\subsection{Alignment Trust}
\label{sec:trust}

The shared groups let us compare how distinctly each modality supports
the assigned class relative to competing classes. This comparison should
reflect relative semantic support without depending on a score row's
overall offset or positive scale. Alignment Trust removes both by
standardizing the scores against all $C$ anchors. For each group mean,
let $\ell_{c,b}^m=\mu_{c,b}^m A^\top$ and define
\begin{equation}
 z_{c,b}^m=
 \frac{\ell_{c,b}^m-\operatorname{mean}_{j}([\ell_{c,b}^m]_j)\mathbf 1}
      {\operatorname{std}_{j}([\ell_{c,b}^m]_j)}.
 \label{eq:standardized_scores}
\end{equation}
Here $j$ ranges over all anchors, $\mathbf 1$ is the all-ones vector,
and $\operatorname{std}$ is the population standard deviation; set
$z_{c,b}^m=0$ for a zero-variance vector.
Dot products and cosine similarities therefore give identical standardized
scores. Averaging the $K=|\mathcal C_b|$ assigned-anchor scores measures
semantic support across the accumulated groups:
\begin{equation}
 \alpha_b^m=\frac1K\sum_{c\in\mathcal C_b}[z_{c,b}^m]_c.
 \label{eq:alignment}
\end{equation}
This semantic alignment score corresponds to the comparison in
Figure~\ref{fig:evidence}b (expanded in the appendix).
A softmax then assigns greater weight to the modality with stronger support:
\begin{equation}
 [w_b^v,w_b^t]=\operatorname{softmax}\big(g[\alpha_b^v,\alpha_b^t]\big).
 \label{eq:trust}
\end{equation}
The inverse temperature $g$ controls reweighting strength; equal alignment
gives equal weights.

Alignment measures support for the shared predicted assignments.
The weights remain one half during
warm-up or with fewer than two positive-mass groups. For $C=2$, each
nonconstant standardized row is $\{+1,-1\}$ regardless of its score gap.
Trust uses this binary signal after warm-up; realignment uses the identity map.

\subsection{Anchor Realignment}
\label{sec:realign}\label{sec:discriminative-fit}

Reweighting alone leaves each branch's feature--anchor relationships
unchanged. Anchor Realignment uses the same semantic reference to adjust
these relationships before fusion. It fits one map from fused means
$\mu_{c,b}^0$ to targets $a_c$, then applies it to both modalities.
The fused means provide a joint fitting signal under the assignments
already used for Trust.

The correction acts only in the joint span of the observed means and
their targets, leaving directions outside this span unchanged. We constrain
the map to be orthogonal to preserve inner products before residual mixing.
Stack the means into $X_b\in\mathbb R^{K\times d}$ and select their
assigned rows from $A$ to form $A_{\mathcal C_b}$. Let the columns of
$Q_b\in\mathbb R^{d\times r}$ form an orthonormal basis for this span,
so $Q_b^\top Q_b=I_r$. The full-space map is
\begin{equation}
 R_b(P)=I+Q_b(P-I_r)Q_b^\top,\qquad P^\top P=I_r.
 \label{eq:low_rank_rotation}
\end{equation}
Here $P$ rotates or reflects features within the span, while the map is
identity outside it.
We first fit the fused means to their targets using orthogonal
Procrustes~\citep{schonemann1966procrustes}, giving a geometric initialization:
\begin{equation}
 P_b^0=\arg\min_{P^\top P=I_r}\|X_bQ_bP-A_{\mathcal C_b}Q_b\|_F^2.
 \label{eq:procrustes_initialization}
\end{equation}
Among nonunique optima, we choose the solution nearest identity
(see the appendix).

A close geometric fit alone does not ensure separation from competing
classes. Moving a feature toward its assigned anchor can also increase
its similarity to other anchors. An increase shared across all class
scores leaves their ranking unchanged. We therefore refine the initialization
with a classification objective that compares each assigned anchor against
all alternatives. Define projected mean directions
$\tilde\mu_c=\nu(\mu_{c,b}^0Q_b)$ and projected classifier anchors
$\tilde A_b=AQ_b$. We leave the projected anchors unnormalized to preserve
their original scores within the span. With residual strength $\beta$
and softmax temperature $T$, the corrected mean and its probabilities are
\begin{equation}
 \hat\mu_c(P)=\nu\big((1-\beta)\tilde\mu_c+\beta\tilde\mu_cP\big),\qquad
 p_{c,b}(P)=\operatorname{softmax}\big(\hat\mu_c(P)\tilde A_b^\top/T\big).
 \label{eq:refinement_scores}
\end{equation}
The residual mix retains part of the original mean direction, and the
softmax compares it with all $C$ classifier anchors.
Group concentration $\rho_c=\|\mu_{c,b}^0\|_2$ weights the cross-entropy objective:
\begin{equation}
 \mathcal L_b(P)=
 \frac{\sum_{c\in\mathcal C_b}\rho_c\,\operatorname{CE}(p_{c,b}(P),c)}
      {\sum_{c\in\mathcal C_b}\rho_c}.
 \label{eq:calibration_ce}
\end{equation}
Here $\operatorname{CE}(p,c)=-\log p_c$. Concentrated groups contribute
more to the fit. Projected descent from the Procrustes initialization
produces $P_b$ and the full map $R_b=R_b(P_b)$.
This mean-based objective remains a proxy under the shared pseudo-labels.
The appendix details the local solver.

\subsection{Prediction and online updates}
\label{sec:online-prediction}

Prediction brings the two operations together: Trust weights the branch
scores produced after Realignment. The adaptation state for each batch
uses only earlier observations; the batch updates the history after prediction.
To retain part of the original representation, mix the normalized input
$u=\nu(z)$ with its transformed counterpart $u_R=\nu(uR_b)$:
\begin{equation}
 \phi_b(z)=\nu\big((1-\beta)u+\beta u_R\big).
 \label{eq:residual_transform}
\end{equation}
The score vector is $q_i^m=\tau\phi_b(u_i^m)A^\top$, with $\tau$ the
checkpoint logit scale. The prediction combines the corrected scores
using the semantic modality weights:
\begin{equation}
 s_i=w_b^v q_i^v+w_b^t q_i^t.
 \label{eq:prediction}
\end{equation}

Orthogonality preserves inner products under $R_b$ itself; residual mixing
and normalization can change pairwise angles
(see the appendix). Every observation is predicted,
including those rejected by the memory gate. During warm-up, $\phi_b=\nu$.

After prediction, raw features update the shared sums under shared
pseudo-labels and weights. Corrected predictions never supply memory labels,
keeping both operations tied to the same raw observation history.
The appendix details the stored statistics and update order.

%% file: figures/method_architecture.tex
\begin{figure}[!ht]
\centering
\includegraphics[width=.75\linewidth]{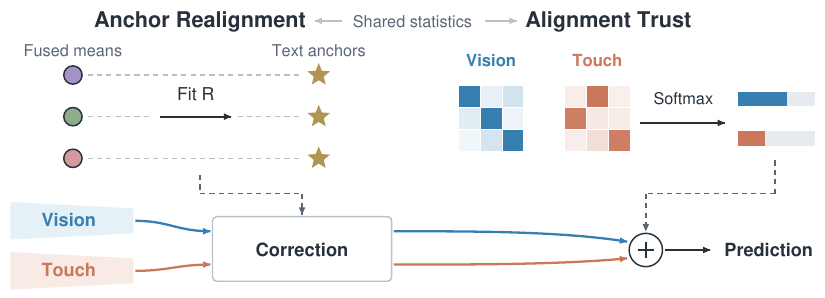}
\caption{\textbf{A shared history supports modality weighting and feature correction.}
Past unlabeled observations supply both modules with shared group statistics.
Anchor Realignment fits $R$ using fused means (circles) and class text
embeddings (stars), then refines class separation.
Alignment Trust compares standardized group--text scores to assign fusion weights.
Dashed arrows pass the shared correction and weights to the prediction path;
the same correction is applied to both branches before weighted fusion.
Encoders remain frozen. Positions, colors and bar lengths are schematic.}
\label{fig:method-architecture}
\end{figure}

%% file: sections/04_related_work.tex
\section{Related Work}
\label{sec:related_work}

\paragraph{Tactile--visual--language representations.}
Shared semantic representations connect contact observations with visual
and linguistic information. Image--text alignment provides a basis for
comparing visual features with class descriptions~\citep{radford2021clip}.
Paired visual--tactile observations establish correspondences between
appearance and contact~\citep{yang2022tag}, while language supervision
connects these observations with semantic descriptions~\citep{fu2024tvl}.
UniTouch and AnyTouch extend representation learning across sensors and
modalities~\citep{yang2024unitouch,feng2025anytouch}. These developments
provide the shared language space in which our adaptation operates.
We study how to use that space during deployment, when the relative
usefulness of the two sensory representations changes. In \method,
class text embeddings serve both as classification references and as
targets for comparing and correcting sensory features.

\paragraph{Test-time adaptation and modality reliability.}
Test-time adaptation learns from unlabeled observations during deployment.
TENT minimizes entropy~\citep{wang2021tent}; SAR combines reliable-sample
selection with sharpness-aware optimization~\citep{niu2023sar}; DeYO filters
and reweights samples using entropy and prediction changes under
perturbation~\citep{lee2024deyo}. RoTTA uses memory and teacher--student
updates~\citep{yuan2023rotta}. These methods derive adaptation signals
from predictions and accumulated observations, with different mechanisms
for controlling unreliable updates.
Multimodal adaptation additionally considers how updates should respond
when branches are affected differently. READ adapts
cross-modal attention~\citep{yang2024read}, SuMi combines sample selection
with information sharing~\citep{guo2025sumi}, and TSA routes modality-specific
adapters~\citep{chen2025tsa}. These approaches make modality interactions
part of the adaptation problem. \method\ addresses their relative
contributions through a semantic comparison: it evaluates accumulated
visual and tactile observations against the same class text embeddings.
Shared observations and pseudo-label groups give this comparison a common
reference across the two modalities.

\paragraph{Adaptation with frozen encoders.}
Adaptation can also operate on cached features, semantic references or
contextual embeddings while keeping encoders frozen.
TDA uses a feature--pseudo-label cache~\citep{karmanov2024tda};
OGA combines Gaussian likelihoods with zero-shot priors~\citep{fuchs2025oga};
StatA uses text-derived statistical anchors~\citep{zanella2025stata}.
ETTA adapts contextual embeddings and prompt selection~\citep{dastmalchi2025etta};
SubTTA aligns visual and textual subspaces~\citep{zeng2026subtta}.
MG-MTTA freezes encoders and adapts fusion using modality drift from
running anchors~\citep{chen2026mgmtta}.
These studies adapt how pretrained features are interpreted, including
their class associations, alignment and contributions to the final prediction.
Our TDA, OGA and StatA extensions adapt each modality separately and average
their outputs. \method\ couples modality weighting and feature correction
through one shared memory. Alignment Trust compares semantic support
under common class assignments, while Anchor Realignment adjusts both
modalities with a common correction. Its classification refinement
compares each assigned text embedding with competing classes, extending
the geometric alignment objective to class discrimination. This coupling
connects the choice of modality contributions with the class distinctions
preserved by the corrected features.

%% file: sections/05_experiments.tex
\Needspace{20\baselineskip}
\section{Experiments}
\label{sec:experiments}
\suppressfloats[t]

\input{sections/02_benchmark}

\subsection{Recognition across sensing conditions}
\label{sec:exp-failed}\label{sec:exp-touch}
\input{generated/sa_layout_main}
\input{figures/experiments}

\paragraph{Recovering recognition when touch is unreliable.}
SenseAlign reduces the loss caused by weak tactile predictions.
Across four UniTouch natural-shift panels, mean accuracy rises from
43.7\% with equal fusion to 56.8\%, exceeding frozen vision's 54.9\%
(Table~\ref{tab:sa-unitouch}). Gains hold in every panel and displayed
TacQuad condition (Figure~\ref{fig:experiments}b).
Default weights favor vision (Figure~\ref{fig:experiments}a);
accuracy reflects both weighting and realignment.

\paragraph{Retaining the contribution of stronger touch.}
The same semantic rule can favor touch when it is more informative.
On AnyTouch OFR, frozen touch outperforms vision (74.6\% versus 37.3\%),
and default weights favor touch (Figure~\ref{fig:experiments}a).
With $g=1$, SenseAlign reaches 81.2\%, above both frozen touch and
71.7\% equal fusion. Across four AnyTouch natural-shift panels,
mean accuracy rises from 45.2\% to 53.8\%; clean TacQuad still trails
TDA (45.5\% versus 46.3\%).

\paragraph{Improving recognition with two informative modalities.}
Gains extend beyond severe modality imbalance.
SenseAlign improves on equal fusion in all six TAG-C corruption scenarios,
reaching 60.9\% mean accuracy versus 53.8\% for equal fusion and
55.6\% for TDA, the strongest complete baseline
(Table~\ref{tab:sa-tagc}).
SenseAlign leads all three AnyTouch scenarios and the UniTouch mean;
SAR leads UniTouch Visual and Mixed.

\input{generated/sa_layout_tagc}

\paragraph{A common configuration retains the mean advantage.}
With $\beta=0.2$, $g=4$ and warm-up 16 in all 14 main-table scenarios,
SenseAlign averages 56.62\%, versus 49.00\% for TDA: a 7.61-point gain.
The mean advantage thus persists across both backbones without
scenario-specific settings (see the appendix).

\subsection{Semantic support beyond confidence}
\label{sec:weighting-controls}

Semantic support improves weighting even when corrected features and
fitted maps are held constant. Trust exceeds maximum-probability,
inverse-entropy and equal weighting by 0.94, 0.80 and 1.22 points on
average across six TAG-C scenarios (see the appendix).
The paired 95\% interval against equal weighting is $[0.89,1.56]$,
conditional on recorded histories. Probability and entropy use current
predictions; Trust uses accumulated observations.

The advantage persists against history-grouped margin, which uses the
same observations, groups and weights but compares top-two score gaps
instead of assigned-class support. Trust retains mean gains of 3.33 points across three UniTouch
TacQuad/RCT panels and 1.26 across six TAG-C panels
(see the appendix). The semantic signal thus adds value beyond history aggregation alone.

\subsection{Feature correction for class discrimination}
\label{sec:realignment-finding}

Refinement improves recognition beyond geometric matching to text
embeddings. With AnyTouch and equal weights, geometric realignment raises
OFR accuracy from 71.7\% to 74.4\%; refinement further raises it to
78.3\%, isolating the gain from Trust. A retrospective one-split diagnostic
applies final-memory maps to their construction observations:
refinement lowers correct-class similarity but lowers the strongest rival
more, increasing their mean gap (see the appendix). This illustrates
improved class separation, separately from online evaluation.

\FloatBarrier
\Needspace{12\baselineskip}
\subsection{Combining weighting and realignment}
\label{sec:ablation}
\input{generated/sa_layout_ablation}

Trust adjusts modality contributions; Realign changes the features used
to score classes. Their combination leads on all three sensors in the
TacQuad component study (Table~\ref{tab:sa-ablation}), reaching 47.99\%
mean accuracy: 3.72 points above Trust alone and 6.25 above Realign alone.
These matched comparisons support complementary roles for weighting
and feature correction.

%% file: sections/02_benchmark.tex
\subsection{Datasets and checkpoints}
\label{sec:benchmark}

We evaluate the released UniTouch~\citep{yang2024unitouch} and
AnyTouch~\citep{feng2025anytouch} checkpoints without retraining.
Natural shifts cover object recognition on TacQuad~\citep{feng2025anytouch}
and material recognition on ObjectFolder Real (OFR)~\citep{gao2023objectfolder}
and RCT~\citep{he2026rct}. Controlled corruptions use TAG-C, based on
Touch and Go~\citep{yang2022tag}, with visual, tactile and mixed corruption
streams. Both models use the same evaluation panels.
Table~\ref{tab:regimes} summarizes representative modality relationships,
stream sizes and frozen accuracies.
The appendix provides the comparison protocol and implementation details.

\input{generated/sa_layout_regimes}

%% file: generated/sa_layout_regimes.tex
\begin{table}[ht]
\centering
\footnotesize
\caption{Evaluation protocol and frozen accuracy (\%), one original prompt per class. Presentations count paired observations. Touch is unreliable near/below chance (4.2\% for 24 classes, 14.3\% for seven), dominant when more accurate; both sensors are informative when well above chance. Labels are retrospective and unavailable to adaptation. Natural shifts weight groups and streams equally; TAG-C aggregation is detailed in the appendix.}
\label{tab:regimes}
\setlength{\tabcolsep}{2.8pt}
\begin{tabular*}{\linewidth}{@{\extracolsep{\fill}}>{\raggedright\arraybackslash}p{1.45cm}llcrcccc@{}}
\toprule
 & & & & & & \multicolumn{3}{c}{Frozen accuracy} \\
\cmidrule(l){7-9}
Regime & Benchmark & Model & Streams & Presentations & Classes & Vision & Touch & Fusion \\
\midrule
 & TacQuad (clean) & UniTouch & 3 sensors & 1,536 & 24 & 54.5 & \phantom{0}0.6 & 46.4 \\
 & TacQuad (15 corruptions) & UniTouch & 45 & 23,040 & 24 & 53.7 & \phantom{0}0.6 & 39.4 \\
 & ObjectFolder Real & UniTouch & 5 splits & 3,395 & 7 & 48.7 & 16.2 & 40.8 \\
\multirow{-4}{=}{Touch\newline unreliable} & RCT & UniTouch & 1 & 9,620 & 7 & 62.8 & 11.4 & 48.1 \\
\midrule
Touch\newline dominant & ObjectFolder Real & AnyTouch & 5 splits & 3,395 & 7 & 37.3 & 74.6 & 71.7 \\
\midrule
 & TAG-C (visual / tactile) & UniTouch & 2 & 90,772 & 20 & 51.3 & 41.2 & 51.6 \\
 & TAG-C (visual / tactile) & AnyTouch & 2 & 90,772 & 20 & 55.8 & 43.5 & 57.6 \\
\multirow{-3}{=}{Both\newline informative} & TAG-C mixed stream & AnyTouch & 1 & 94,898 & 20 & 52.0 & 42.4 & 55.2 \\
\bottomrule
\end{tabular*}
\end{table}

%% file: generated/sa_layout_main.tex
\begin{table}[t]
\centering
\footnotesize
\caption{Natural-shift accuracy (\%) with one prompt per class. Both models share four panels; Mean weights panels equally. OFR uses five released-order splits. Frozen$^{\circ}$ uses equal fusion.}
\label{tab:sa-unitouch}
\setlength{\tabcolsep}{3pt}
\begin{tabular}{@{}lrrrrr@{\hspace{8pt}}rrrrr@{}}
\toprule
 & \multicolumn{5}{c}{UniTouch} & \multicolumn{5}{c}{AnyTouch} \\
\cmidrule(lr){2-6}\cmidrule(l){7-11}
Method & \shortstack{TacQuad\\clean} & \shortstack{TacQuad\\corr.} & OFR & RCT & Mean & \shortstack{TacQuad\\clean} & \shortstack{TacQuad\\corr.} & OFR & RCT & Mean \\
\midrule
Frozen$^{\circ}$ & 46.4 & 39.4 & 40.8 & 48.1 & 43.7 & 42.2 & \underline{37.4} & 71.7 & 29.5 & \underline{45.2} \\
Frozen Vision & \underline{54.5} & \underline{53.7} & \underline{48.7} & \underline{62.8} & \underline{54.9} & 42.2 & 32.0 & 37.3 & \underline{47.1} & 39.6 \\
Frozen Tactile & 0.6 & 0.6 & 16.2 & 11.4 & 7.2 & 18.1 & 18.1 & \underline{74.6} & 18.1 & 32.3 \\
\midrule
TENT & 45.5 & 36.8 & 35.0 & 26.5 & 35.9 & 41.1 & 35.1 & 69.7 & 19.4 & 41.3 \\
SAR & 46.4 & 39.3 & 40.6 & 39.7 & 41.5 & 42.0 & 37.2 & 71.6 & 24.7 & 43.9 \\
DeYO & 46.4 & 39.4 & 40.7 & 46.5 & 43.2 & 42.2 & 37.2 & 71.8 & 28.2 & 44.9 \\
SuMi & 46.4 & 38.9 & 39.3 & 26.2 & 37.7 & 42.2 & 37.4 & 71.9 & 18.0 & 42.4 \\
RoTTA & 40.0 & 34.5 & 26.8 & 26.3 & 31.9 & 40.8 & 34.9 & 67.8 & 3.2 & 36.7 \\
TSA & 43.5 & 36.3 & 37.6 & 11.9 & 32.3 & 44.4 & 32.8 & 59.6 & 17.8 & 38.6 \\
TDA & 49.1 & 43.6 & 42.6 & 37.2 & 43.1 & \textbf{46.3} & 37.1 & 69.3 & 27.5 & 45.0 \\
OGA & 52.1 & 46.1 & 48.1 & 9.1 & 38.8 & 39.9 & 32.8 & 68.1 & 19.9 & 40.2 \\
StatA & 51.3 & 45.2 & 46.2 & 60.2 & 50.7 & 30.1 & 25.7 & 55.3 & 37.3 & 37.1 \\
\midrule
\rowcolor{oursrow}SenseAlign & \textbf{58.6} & \textbf{54.0} & \textbf{50.4} & \textbf{64.1} & \textbf{56.8} & \underline{45.5} & \textbf{39.0} & \textbf{81.2} & \textbf{49.6} & \textbf{53.8} \\
\bottomrule
\end{tabular}
\end{table}

%% file: figures/experiments.tex
\begin{figure}[t]
\centering
\includegraphics[width=.92\linewidth]{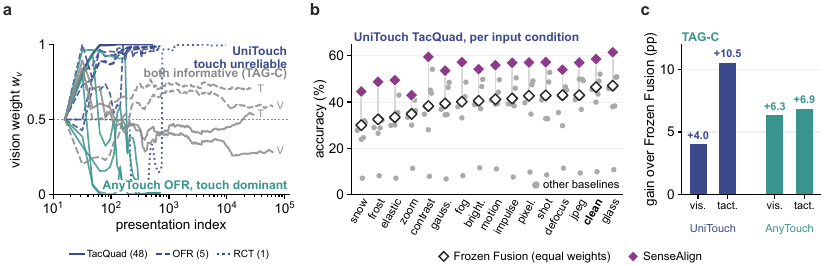}
\caption{\textbf{Modality weights adapt across streams, with gains over equal fusion.}
(a) Online vision weights: the median across TacQuad streams and individual
trajectories elsewhere. Grey TAG-C curves are solid for AnyTouch and dashed
for UniTouch; V/T indicate visual/tactile corruption.
(b) UniTouch TacQuad accuracy averaged over three sensors per condition.
Hollow diamonds denote frozen equal fusion, filled diamonds \method,
and grey dots other baselines.
(c) TAG-C gains over matched equal fusion in percentage points under visual
or tactile corruption; blue denotes UniTouch and teal AnyTouch.}
\label{fig:experiments}
\end{figure}

%% file: generated/sa_layout_tagc.tex
\begin{table}[t]
\centering
\fontsize{8.5}{10}\selectfont
\caption{TAG-C severity-3 accuracy (\%), one prompt per class; means weight scenarios equally. $^{\circ}$: equal fusion. Bold/underline: best/second. }
\label{tab:sa-tagc}\label{tab:sa-anytouch-tagc}\label{tab:sa-unitouch-tagc}
\setlength{\tabcolsep}{3pt}
\begin{tabular}{@{}lccccccccc@{}}
\toprule
& \multicolumn{4}{c}{AnyTouch} & \multicolumn{4}{c}{UniTouch} & Overall \\
\cmidrule(lr){2-5}\cmidrule(lr){6-9}\cmidrule(l){10-10}
Method & Visual & Tactile & Mixed & Mean & Visual & Tactile & Mixed & Mean & Mean \\
\midrule
Frozen$^{\circ}$ & 56.1 & 59.1 & 55.2 & 56.8 & 52.8 & 50.3 & 49.6 & 50.9 & 53.8 \\
TENT & 53.0 & 60.2 & 2.4 & 38.6 & 46.3 & 32.2 & 33.6 & 37.4 & 38.0 \\
SAR & 52.4 & 56.9 & 44.1 & 51.1 & \textbf{58.9} & 53.1 & \textbf{57.7} & \underline{56.6} & 53.9 \\
SuMi & 56.7 & 60.8 & 51.6 & 56.4 & 49.8 & 45.4 & 16.0 & 37.1 & 46.7 \\
RoTTA & \underline{62.2} & 57.1 & 50.0 & 56.4 & 42.4 & 47.4 & 31.3 & 40.4 & 48.4 \\
DeYO & 58.0 & 59.5 & 57.2 & 58.2 & 53.4 & 51.4 & 49.8 & 51.5 & 54.9 \\
TSA & 37.0 & 41.9 & 30.5 & 36.5 & 35.2 & 37.6 & 32.8 & 35.2 & 35.9 \\
TDA & 57.2 & \underline{62.1} & \underline{57.5} & \underline{58.9} & 54.0 & 51.7 & 50.8 & 52.2 & \underline{55.6} \\
OGA & 46.9 & 47.1 & 49.2 & 47.7 & 40.0 & 48.0 & 37.8 & 41.9 & 44.8 \\
StatA & 50.5 & 56.4 & 50.6 & 52.5 & 52.6 & \underline{53.7} & 50.7 & 52.3 & 52.4 \\
\rowcolor{oursrow}SenseAlign & \textbf{62.4} & \textbf{66.0} & \textbf{61.4} & \textbf{63.3} & \underline{57.7} & \textbf{60.8} & \underline{57.1} & \textbf{58.5} & \textbf{60.9} \\
\bottomrule
\end{tabular}
\end{table}

%% file: generated/sa_layout_ablation.tex
\begin{wraptable}{R}{0.5\textwidth}
\centering\footnotesize
\captionsetup{width=\linewidth,font=footnotesize,justification=raggedright,singlelinecheck=false}
\caption{\textbf{Components on AnyTouch TacQuad.} One prompt per class; accuracy (\%). Groups are weighted equally within each sensor; Mean weights sensors equally.}
\label{tab:sa-ablation}
\setlength{\tabcolsep}{2pt}
\begin{tabular}{@{}lrrrr@{}}
\toprule
Configuration & \shortstack{GelSight\\Mini} & DIGIT & DuraGel & Mean \\
\midrule
Equal fusion & 55.15 & 42.31 & 30.23 & 42.56 \\
Trust only & 46.76 & 45.06 & 41.01 & 44.27 \\
Realign only & 54.18 & 38.34 & 32.69 & 41.74 \\
SenseAlign & \textbf{56.50} & \textbf{46.23} & \textbf{41.24} & \textbf{47.99} \\
\bottomrule
\end{tabular}
\end{wraptable}

%% file: sections/06_conclusion.tex
\section{Conclusion}
\label{sec:conclusion}

\method\ uses one unlabeled history to compare vision and touch against
text semantics and fit a shared classification correction.
With frozen encoders, it improves recognition across natural shifts and
controlled corruptions under the evaluated protocols. Matched TAG-C
controls show that semantic weighting adds an average benefit beyond
realignment and simple confidence-based fusion. The relative benefits depend on
the backbone and sensing condition.

%% file: appendix/protocol_details.tex
\section{Evaluation protocol in detail}
\label{app:protocol}

Table~\ref{tab:regimes} summarizes representative modality relationships,
stream sizes and frozen accuracies.

\paragraph{Common comparison settings.}
Within each comparison, all methods share the checkpoint, ordered inputs,
candidate vocabulary and one original prompt per class. Comparisons use
nominal external batches of 16 while retaining each baseline's native
internal steps and serving timing. SenseAlign predicts before updating
and shares its text anchors between classification and correction.
Encoder paths, baseline adaptations and task-specific settings are given
in Appendices~\ref{app:evaluation}, \ref{app:unitouch-baselines}
and~\ref{app:hyperparameters}.

\paragraph{Two pretrained models.}
UniTouch~\citep{yang2024unitouch} and AnyTouch~\citep{feng2025anytouch}
use the same four natural-shift panels and the controlled-corruption
streams. Both are released
checkpoints; we do not retrain them.

\paragraph{Natural sensor shifts.}
TacQuad~\citep{feng2025anytouch} provides object recognition across
GelSight Mini, DIGIT and DuraGel sensors. We keep a fixed 512 observations
per sensor, selected by path hash. We evaluate clean inputs and 15 visual
corruptions while holding tactile inputs fixed, giving 48 streams with
24 classes. ObjectFolder
Real (OFR)~\citep{gao2023objectfolder} provides seven-class material recognition
from real tactile images and object views. We use all five source splits in
their original order: 3,395 presentations from 61 objects.
RCT~\citep{he2026rct} contributes the full seven-category DIGIT material
stream, with 9,620 observations from 121 materials. 
The AnyTouch extension in Table~\ref{tab:sa-unitouch} uses these same
sample identities, class vocabularies, stream orders and external batches
of 16. Each sensor/condition stream starts with fresh adaptation state;

\paragraph{Controlled corruptions.}
TAG-C, a controlled-corruption benchmark based on Touch and Go
(TAG)~\citep{yang2022tag}, presents 4,126 frame
identities under 15 visual and seven tactile corruptions at severity
three. The visual and tactile streams hold 61,890 and 28,882 presentations
with 20 material classes, each run as one continuous stream in its
original order. With both backbones we additionally run the same mixed stream of 94,898
presentations in which clean inputs and visual, tactile and joint
corruptions are interleaved.

\paragraph{Prompt configuration.}
Every method uses one original prompt per class on every benchmark.
Within each comparison, SenseAlign and the baselines share the ordered
class vocabulary and text prompts.

\paragraph{Compared methods.}
We evaluate baselines from three test-time adaptation families:
parameter-update methods (TENT, SAR, DeYO, RoTTA, SuMi), readout-adaptation
methods (TDA, OGA, StatA) and multimodal-reliability methods (READ, TSA).
Each keeps its native internal steps and output timing; TDA, OGA and StatA
maintain one state per sensor and average the two. Appendix~\ref{app:unitouch-baselines}
details the baselines and their precision qualifications. For \method\ we
report the two modules separately and together.

\paragraph{Metrics.}
The four natural-shift panels are TacQuad clean, TacQuad corruptions,
ObjectFolder Real and RCT. Object or material groups receive equal weight
within each stream, followed by equal weight over streams. Each natural-shift mean gives
equal weight to its four panels. AnyTouch TAG-C
accuracy weights presentations; UniTouch averages corruption conditions
within each direction (Appendix~\ref{app:unitouch-tagc}). Mixed accuracy
weights presentations for both backbones. Each backbone mean weights its
three scenarios equally; Overall weights all six equally, using unrounded
scores and requiring every constituent result. On the AnyTouch TAG-C
streams our runs take batches of 16 with a
two-observation tail.

\paragraph{Motivation figure protocol.}
\label{app:motivation-figures}
Figure~\ref{fig:regimes} uses single prompts and the $g=4$ replay;
its ObjectFolder pairs are retrospective cases selected from the first
test split by the largest loss from equal fusion.
Figure~\ref{fig:evidence} uses the two ObjectFolder streams and single
prompts from the first two rows of Figure~\ref{fig:regimes}(a),
including all seven classes rather than only the selected pairs. Both models
use the same 704 observations. Its matrices compare per-sensor class
means with all semantic anchors at the end of the stream. The scalar
$\alpha$ averages the standardized assigned-anchor scores as in
Eq.~\ref{eq:alignment}; the displayed mean vision weight uses $g=4$.

\paragraph{Experiment figure protocol.}
Figure~\ref{fig:experiments} uses the $g=4$ replay on recorded
streams. Panel (a)
contains 63 single-prompt histories: 48 for TacQuad, five UniTouch OFR
splits, one RCT stream, five AnyTouch OFR splits, and four TAG-C streams
(two corruption directions per model).
TacQuad is summarized by its median trajectory; the other streams are
shown individually. Panel (b) averages accuracy over the three
sensors for each condition. Its nine baseline series are TENT, SAR,
DeYO, SuMi, RoTTA, READ, TSA, TDA and OGA.
Panel (c) reports percentage-point gains over the same-head equal fusion
under visual or tactile corruption, with one prompt per class for both models.

%% file: appendix/tact_method_details.tex
\section{Method specification and reproducibility}
\label{app:method_details}

\paragraph{Terminology shared with the main text.}
Alignment Trust (Trust) performs modality weighting; Anchor Realignment
(Realign) performs feature realignment, including classification refinement.
\method\ combines both modules using one shared memory.
``Trust only'' and ``Realign only'' in Table~\ref{tab:sa-ablation}
denote the individual modules.
Narrow appendix tables use ``Realign'' for feature realignment only and
``Realign$^{-}$'' for realignment without classification refinement.
Both realignment variants retain equal modality weights.
Equal-weight fusion, shortened to ``equal fusion'', uses neither module;
its prompt configuration is specified for each comparison.

\subsection{Semantic anchors and memory selection}
\label{app:anchor-selection}

One original prompt per class supplies the normalized text anchor $a_c$
(Appendix~\ref{app:prompts}). Its embedding is used for classification,
modality comparison and the realignment target. Equation~\ref{eq:memory_selection}
defines the acceptance indicator $\gamma_i$, pseudo-label $\widehat y_i$
and margin weight $\lambda_i$. The top-class score and runner-up margin use unscaled
cosine similarities. All three are computed from the raw equal-fusion
feature before any realignment or modality reweighting.

\subsection{State and execution order}
\label{app:method_state}

Let $I_{<b}$ contain all observations before batch $b$.
To accumulate the means in Equation~\ref{eq:class_means}, we store
\begin{equation}
 S_{c,b}^m=\sum_{j\in I_{<b}}\mathbf 1[\widehat y_j=c]\lambda_j u_j^m,
 \qquad
 \Lambda_{c,b}=\sum_{j\in I_{<b}}\mathbf 1[\widehat y_j=c]\lambda_j.
 \label{eq:shared_sums}
\end{equation}
Dividing each feature sum by its shared total weight gives
Equation~\ref{eq:class_means} in the main text.
The following realignment updates suppress the batch index and write
$S_c=S_c^0$ and $\Lambda_c$ for its fused-feature sum and shared mass.
The per-modality sums use the same assignments and weights.

Anchor Realignment is identified by
\texttt{mcc\_readout\_ce} in the source records. \method\ additionally keeps
the per-modality sums of Equation~\ref{eq:class_means} for Alignment Trust.
Anchor Realignment maintains a feature sum
$S_c\in\mathbb R^d$ and mass $\Lambda_c$ for every class, an accepted-observation
count $N$, and a reusable factorization of the current correction. The
encoders and semantic anchors are fixed. After predicting batch $b$, it applies
\begin{align}
 S_c&\leftarrow S_c+\sum_{i\in I_b}\mathbf 1[\widehat y_i=c]\,\lambda_i\,u_i^0,\nonumber\\
 \Lambda_c&\leftarrow \Lambda_c+\sum_{i\in I_b}\mathbf 1[\widehat y_i=c]\,\lambda_i,\qquad
 N\leftarrow N+\sum_{i\in I_b}\gamma_i,
 \label{eq:appendix_accumulator}
\end{align}
with $\widehat y_i$, $\lambda_i$ and $\gamma_i$ as in Equations~\ref{eq:memory_selection}
and~\ref{eq:class_means}.
A mean $S_c/\Lambda_c$ is available when $\Lambda_c>0$. Warm-up counts 16 accepted
observations, not a weight mass of 16 and not a fixed number of batches.
The gate uses strictly positive unscaled cosine support and top-two margin.
A tie therefore rejects the observation for updating the statistics,
although it still receives a classification prediction. No forgetting factor
or sample replay buffer is used.

\begin{table}[!ht]
\centering\small
\caption{Anchor Realignment alone for one batch. All fitted state is based on earlier batches; fusion weights remain equal. \method\ also computes modality weights from the shared memory, as specified in Section~\ref{sec:trust}.}
\label{tab:tact-algorithm}
\begin{tabular}{p{.06\linewidth}p{.86\linewidth}}
\toprule
Step & Operation \\
\midrule
1 & Encode both modalities with frozen encoders; normalize their features. \\
2 & If $C>2$, $N\geq16$ and history changed, form $\mu_c=S_c/\Lambda_c$, construct $Q$, initialize $P$ by minimum-change Procrustes, and refine $P$ using Equation~\ref{eq:calibration_ce}. \\
3 & Apply the correction fitted to earlier observations to each branch, fuse with equal weights, classify with the shared text anchors, and record every prediction. Use the identity map before warm-up or when $C=2$. \\
4 & Compute the pseudo-label and positive-margin weight from raw frozen fusion. \\
5 & Append the current batch's raw statistics using Equation~\ref{eq:appendix_accumulator}. Mark the fit for recomputation if any observation was accepted. \\
\bottomrule
\end{tabular}
\end{table}

True labels, corrupted-modality
identities and shift boundaries do not enter any operation above.
Adaptation states reset between the independently defined streams, not
at unannounced changes inside a mixed stream.

\subsection{Expanded semantic alignment}
\label{app:alignment-expanded}

Let $M_b^m[c,\cdot]=\mu_{c,b}^m A^\top$ collect scores against all $C$
classifier anchors for the $K=|\mathcal C_b|$ represented groups.
The z-score notation in Equation~\ref{eq:alignment} expands to
\begin{equation}
 \alpha_b^m=\frac1K\sum_{c\in\mathcal C_b}
 \frac{M_b^m[c,c]-\operatorname{mean}_{c'}M_b^m[c,c']}
      {\operatorname{std}_{c'}M_b^m[c,c']},
 \label{eq:alignment_expanded}
\end{equation}
The standard deviation is the population standard deviation; a
zero-variance row contributes zero. The softmax in Equation~\ref{eq:trust}
is equivalently
\begin{equation}
 w_b^m=\frac{\exp(g\alpha_b^m)}{\exp(g\alpha_b^v)+\exp(g\alpha_b^t)}.
 \label{eq:trust_expanded}
\end{equation}

\subsection{Initialization in the joint semantic subspace}
\label{app:procrustes}

Equations~\ref{eq:low_rank_rotation} and~\ref{eq:procrustes_initialization}
define the full-space map and its initialization. We give the
minimum-change solution below.

We drop the batch index of Section~\ref{sec:realign}. For the $K$ means $X$
and assigned anchor rows $A_{\mathcal C}$, the Procrustes objective is
\begin{equation}
 \|XR-A_{\mathcal C}\|_F^2=\|X\|_F^2+\|A_{\mathcal C}\|_F^2-2\langle R,X^\top A_{\mathcal C}\rangle_F.
\end{equation}
The joint row span of $X$ and $A_{\mathcal C}$ has an orthonormal basis $Q$ of rank
$r\leq\min(d,2K)$. Set $Y=(XQ)^\top(A_{\mathcal C}Q)$ and write its nonzero singular
part as $U_+\Sigma_+V_+^\top$. A primary optimum in the joint span has form
\begin{equation}
 P^0=U_+V_+^\top+U_0HV_0^\top,\qquad H^\top H=I.
\end{equation}
Here $U_0,V_0$ span the respective null spaces. The completion $H$ maximizes
$\langle H,U_0^\top V_0\rangle_F$, equivalently selecting the primary fit
closest to identity. A polar factor supplies this completion. Remaining
ties are resolved by pairing canonical null-space bases. The action outside
the joint span is identity. This is the solution for realignment without
refinement and the initialization of Anchor Realignment.

A canonical basis is constructed greedily from the residual subspace
projector. At each step, the coordinate with largest projected squared norm
is selected, with first-coordinate tie breaking. Its vector is twice
reorthogonalized against previous vectors, then normalized and assigned a
positive pivot sign.
Rank decisions use float64 and the threshold
$\max(n_1,n_2)\epsilon_{64}\sigma_1$ for a matrix of shape $(n_1,n_2)$.
These conventions specify otherwise ambiguous rank-deficient solutions.

\subsection{Classification refinement and local solver}

\label{app:refinement}

Define the projected unit mean and unnormalized projected anchors by
$\tilde\mu_c=\nu(\mu_{c,b}^0Q_b)$ and $\tilde a_j=a_jQ_b$.
The vector softmax in Equation~\ref{eq:refinement_scores} has components
\begin{equation}
 [p_{c,b}(P)]_j=
 \frac{\exp(\hat\mu_c(P)\tilde a_j^\top/T)}
      {\sum_k\exp(\hat\mu_c(P)\tilde a_k^\top/T)}.
 \label{eq:refinement_probabilities}
\end{equation}
The residual mean and expanded loss are
\begin{equation}
 \begin{aligned}
 \mathcal L_b(P)&=\frac{\sum_c\rho_c\,\ell_c(P)}{\sum_c\rho_c},\qquad
 \hat\mu_c(P)=\nu\big((1-\beta)\tilde\mu_c+\beta\tilde\mu_cP\big),\\
 \ell_c(P)&=-\log\frac{\exp(\hat\mu_c(P)\tilde a_c^\top/T)}{\sum_j\exp(\hat\mu_c(P)\tilde a_j^\top/T)}.
 \end{aligned}
 \label{eq:calibration_ce_expanded}
\end{equation}
All sums over $c$ use $\mathcal C_b$, and each softmax spans all
vocabulary anchors. The abbreviations in Equation~\ref{eq:residual_transform}
also preserve the full nested normalization:
\begin{equation}
 \phi_b(z)=\nu\big((1-\beta)\nu(z)+\beta\,\nu(\nu(z)R_b(P_b))\big).
 \label{eq:residual_transform_expanded}
\end{equation}
The vector prediction in Equation~\ref{eq:prediction} has components
\begin{equation}
 q_{ic}^m=\tau\phi_b(u_i^m)a_c^\top,\qquad
 s_{ic}=w_b^vq_{ic}^v+w_b^tq_{ic}^t.
 \label{eq:prediction_expanded}
\end{equation}

The proxy in Equation~\ref{eq:calibration_ce} uses unit mean directions in
the joint subspace but retains the original mean norm $\rho_c$ as a class
weight. Since means are averages of unit vectors, this norm represents
concentration rather than sample count. Every class with positive mass
contributes one term; all vocabulary anchors appear in its softmax
normalizer. The projected classifier vectors $\tilde a_j=a_jQ$ are not
renormalized.

For matrix $P$, let $G=\nabla_P\mathcal L_b(P)$. A trial step is
\begin{equation}
 P'=\operatorname{polar}\!\left(P-\eta G/\|G\|_F\right).
\end{equation}
For a matrix with singular value decomposition $U\Sigma V^\top$, the
implemented polar projection is $UV^\top$.
It permits either determinant sign and imposes no determinant-$+1$
constraint. Starting with $\eta=1$, backtracking accepts a candidate only if
\begin{equation}
 \langle G,P'-P\rangle_F<0,\qquad
 \mathcal L_b(P')\leq\mathcal L_b(P)
 +10^{-4}\langle G,P'-P\rangle_F.
\end{equation}
Otherwise $\eta$ is halved, for at most 20 attempts. The solver stops if no
descent step is accepted, $\|G\|_F<10^{-10}$, the accepted decrease is at
most $10^{-9}\max(1,|\mathcal L_b(P)|)$, or 64 accepted steps
have been taken. Gradients are computed with respect to $P$ only; no
backpropagation through the polar decomposition or the encoders is used.
Each new fit restarts from the geometric initialization for the current
history. It does not warm-start from the previous fitted $P$.

\subsection{Hyperparameter settings}
\label{app:hyperparameters}

\begin{table}[!ht]
\centering\small
\caption{Default SenseAlign settings. Main-table exceptions are listed in Table~\ref{tab:selected-settings}. All comparisons use one original prompt per class.}
\label{tab:method_hyperparameters}
\begin{tabular}{ll}
\toprule
Setting & Value \\
\midrule
Encoder parameters updated & None \\
Classification reference & One normalized text embedding per class \\
Initialization targets & Same anchors as the classifier \\
History weighting & Positive top-two cosine margin \\
Refinement class weighting & Norm of accumulated class mean \\
Default residual coefficient & $\beta=0.2$ \\
Default Trust inverse temperature & $g=4$ \\
Binary vocabulary & Trust enabled; realignment disabled \\
Warm-up & 16 accepted observations \\
Refinement temperature & $T=0.07$ \\
Local solver budget & 64 projected steps \\
Initial normalized step & 1.0, with backtracking \\
Nominal external batch & 16; original tails retained \\
Within-stream forgetting & None \\
Embedding dimension & 1,024 (UniTouch); 768 (AnyTouch) \\
\bottomrule
\end{tabular}
\end{table}

\begin{table}[!ht]
\centering\small
\caption{SenseAlign settings for every main-table scenario. All rows use
16 accepted observations before adaptation.}
\label{tab:selected-settings}
\begin{tabular}{llrr}
\toprule
Model & Scenario & $\beta$ & $g$ \\
\midrule
AnyTouch & TacQuad clean & 0.2 & 8 \\
AnyTouch & TacQuad corrupted & 0.4 & 1 \\
AnyTouch & ObjectFolder Real & 0.2 & 1 \\
AnyTouch & RCT & 0.1 & 64 \\
AnyTouch & TAG-C Visual & 0.2 & 4 \\
AnyTouch & TAG-C Tactile & 0.2 & 4 \\
AnyTouch & TAG-C Mixed & 0.2 & 4 \\
\midrule
UniTouch & TacQuad clean & 0.2 & 4 \\
UniTouch & TacQuad corrupted & 0.2 & 4 \\
UniTouch & ObjectFolder Real & 0.1 & 64 \\
UniTouch & RCT & 0.2 & 4 \\
UniTouch & TAG-C Visual & 0.4 & 8 \\
UniTouch & TAG-C Tactile & 0.2 & 4 \\
UniTouch & TAG-C Mixed & 0.4 & 8 \\
\bottomrule
\end{tabular}
\end{table}

\input{appendix/hyperparameter_sensitivity}

\subsection{Text construction and vocabulary}
\label{app:prompts}

Each class uses one original prompt, encoded with the frozen text encoder
and normalized to obtain $a_c$. TAG-C uses \emph{the tactile sensor is
touching a \{name\}}. UniTouch and AnyTouch encode the same strings with
their respective text encoders. Each comparison shares its ordered class
vocabulary and prompt strings across methods.

All prompt strings and ordered candidate lists are provided with the
source evidence specifications. ObjectFolder Real candidates are wood, steel,
polycarbonate, plastic, iron, ceramic and glass. RCT candidates are crafts,
metal, paper cardboard, plastic rubber, small items, textiles leather, and
wood bamboo cork. TacQuad uses its 24 capture-derived object candidates.
No prompt text is updated from test labels or optimized during the stream.

\subsection{Storage and computation}

Class sums and masses require $O(Cd)$ storage. Constructing the joint span
costs $O(dK^2+K^3)$ for $K$ represented pseudo-classes. The factored map
stores $O(dr+r^2)$ scalars. Each local step evaluates scores for all $C$
classes and projects an $r\times r$ matrix, with costs of order
$O(Kr^2+KCr+r^3)$. Applying the map to a batch of $n$ observations costs
$O(ndr+nr^2)$. Its main resource saving
is the absence of encoder gradients, student activations and Adam moments.

%% file: appendix/hyperparameter_sensitivity.tex
\subsection{Parameter sensitivity and a common configuration}
\label{app:parameter-sensitivity}

\begin{figure}[!ht]
\centering
\includegraphics[width=\linewidth]{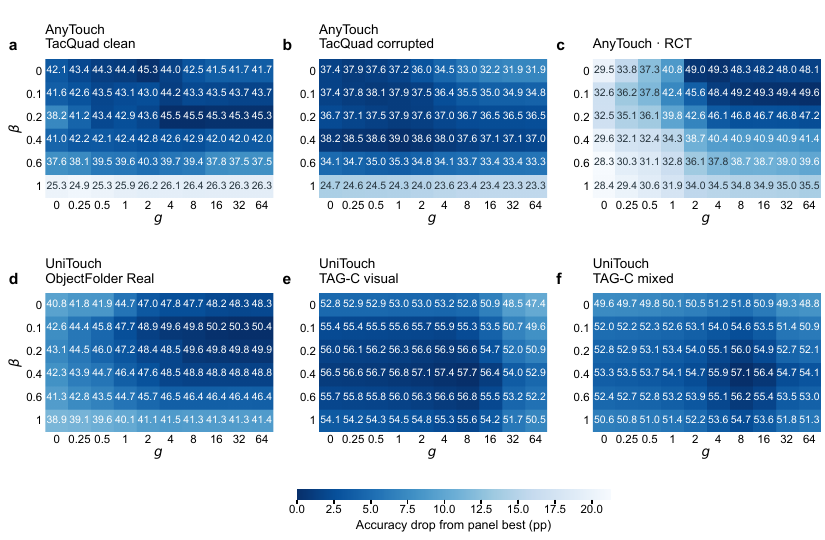}
\caption{\textbf{Sensitivity to feature correction and modality weighting.}
Cells report accuracy (\%); colours show the drop from each panel's best
tested setting on a shared scale.}
\label{fig:beta-g-sensitivity}
\end{figure}

The best-performing settings balance how much to change the features
against how strongly to favor one modality (Figure~\ref{fig:beta-g-sensitivity}).
For AnyTouch RCT and UniTouch OFR, $(\beta,g)=(0.1,64)$ combines a small
feature correction with sharply differentiated modality weights. Increasing
$\beta$ generally reduces accuracy near this setting, suggesting that
redistributing modality contributions is more useful here than strongly
altering the representations. UniTouch TAG-C Visual and Mixed instead
peak at $(0.4,8)$. Their improvement from $\beta=0$ to $0.4$ at $g=8$
supports a larger role for feature correction, while the decline beyond
$g=8$ favors retaining contributions from both modalities.

AnyTouch TacQuad corrupted favors $(\beta,g)=(0.4,1)$, pairing stronger
correction with softer weighting. This pattern is consistent with correcting
features while avoiding an overly sharp preference between modalities.
Clean TacQuad favors $(0.2,8)$, but $g=4$ performs almost identically
at the same $\beta$. Here, moderate correction matters more than the
precise weighting sharpness.

With one configuration ($\beta=0.2$, $g=4$, warm-up 16), SenseAlign
averages 56.62\% across all 14 main-table scenarios, versus 49.00\% for
TDA, the strongest complete baseline, a 7.61-point gain. The mean weights all scenarios equally.

%% file: appendix/tact_evaluation.tex
\section{Primary evaluation with UniTouch}
\label{app:evaluation}

\subsection{Checkpoint and data coverage}

The primary experiments strictly load the vision, touch and text branches
of the released UniTouch checkpoint: 1,030 active tensors with
1,291,745,579 elements and 1,024-dimensional outputs.
The implementation restores text end-token pooling and projection as
specified by the checkpoint and uses the original ImageBind image
preprocessing for both sensory modalities. A fixed official code revision
and weight revision are recorded in the source bundle. CPU/GPU implementation checks
and active-parameter shape checks precede evaluation. The comparison
uses the released encoder with a shared fusion head across methods.

UniTouch's published training data include Touch and Go, Feeling of
Success, YCB-Slide and objects 101--1000 of ObjectFolder 2.0.
The completed TacQuad, ObjectFolder Real and RCT panel is distinct from
those collections. TAG-C is evaluated as corruption robustness on a
documented training source. ObjectFolder
Real is listed as an out-of-domain evaluation in the original work;
TacQuad and RCT are absent from its published training list~\citep{yang2024unitouch}.
This distinction is defined at the dataset level from the published
training list; sample-level exposure remains unspecified because the
released checkpoint lacks a complete sample-level training manifest.

TacQuad retains 512 rows per sensor selected by a fixed path hash before
running the conditions. The same rows appear in the clean condition and
all 15 existing visual corruptions. The original relative order is kept,
and tactile inputs are unchanged across those conditions. There are
25 conservative named-target groups and 24 classes. These names define
the resampling units. TacQuad uses the existing corruption assets without
assigning an official severity level; the severity-three specification
applies to TAG-C. Each of the 48 conditions starts afresh.

ObjectFolder Real retains all five source test splits and their batch tails,
with 3,395 presentations and 2,079 unique image pairs from 61 objects.
Objects may recur across splits. Only the original order is evaluated in
this primary experiment. RCT retains the complete seven-class DIGIT stream
of 9,620 rows and 121 materials. Its source material-identity correction,
image bindings and literal category labels are retained in the data records.
The reference RGB photos are reused across tactile contacts. This is a
material-recognition task distinct from the physical-attribute task in the
AnyTouch supplement.

\subsection{Scoring}

Within each condition, accuracy first averages predictions for an observed
group and then averages the observed groups equally. Conditions are averaged
equally for the task-level result. TacQuad corruption statistics average
45 sensor--corruption conditions; clean accuracy averages three sensors.
ObjectFolder Real averages the five tests, and RCT averages its materials.

Correction and harm counts describe presentations: a correction is a frozen
error changed to a correct prediction, and a harm is the reverse. Their
net divided by presentation count gives the presentation-weighted change;
the primary metric instead gives equal weight to groups.

\subsection{All TacQuad conditions}

Table~\ref{tab:tact-tacquad-conditions} averages sensors for each corruption;
the following tables retain every sensor-specific condition. \method\ is
above equal fusion by point estimate in all 48 sensor--condition streams,
by 2.8 to 31.6 points.

\input{generated/ar_tacquad_conditions}
\FloatBarrier
Tables~\ref{tab:tact-unitouch-tacquad-gelsight}
to~\ref{tab:tact-unitouch-tacquad-duragel} keep each sensor separate.

\input{generated/ar_tacquad_gelsight}
\FloatBarrier
\input{generated/ar_tacquad_digit}
\FloatBarrier
\input{generated/ar_tacquad_duragel}
\FloatBarrier

\subsection{All ObjectFolder Real tests}

\input{generated/ar_ofr_splits}
\FloatBarrier

%% file: generated/ar_tacquad_conditions.tex
\begin{table}[htbp]
\centering\small
\setlength{\tabcolsep}{3.5pt}
\caption{UniTouch on each TacQuad condition (accuracy, \%), averaging named targets within a sensor and then the three sensors. The last column is \method\ minus equal fusion.}
\label{tab:tact-tacquad-conditions}
\begin{tabular*}{\linewidth}{@{\extracolsep{\fill}}lrrr@{}}
\toprule
Condition & Equal fusion & \method & \method\ $-$ fusion \\
\midrule
Clean & 46.45 & 58.59 & +12.14 \\
Brightness & 40.55 & 54.24 & +13.70 \\
Contrast & 38.29 & 59.53 & +21.24 \\
Defocus blur & 42.99 & 53.97 & +10.98 \\
Elastic & 33.43 & 49.52 & +16.09 \\
Fog & 40.16 & 57.26 & +17.10 \\
Frost & 32.54 & 48.77 & +16.24 \\
Gaussian noise & 39.37 & 53.63 & +14.25 \\
Glass blur & 47.09 & 61.54 & +14.45 \\
Impulse noise & 41.57 & 56.99 & +15.42 \\
JPEG & 43.00 & 57.07 & +14.07 \\
Motion blur & 41.21 & 55.99 & +14.78 \\
Pixelation & 42.72 & 57.13 & +14.41 \\
Shot noise & 42.88 & 57.28 & +14.40 \\
Snow & 30.03 & 44.58 & +14.55 \\
Zoom blur & 34.89 & 42.98 & +8.09 \\
\bottomrule
\end{tabular*}
\end{table}

%% file: generated/ar_tacquad_gelsight.tex
\begin{table}[htbp]
\centering\small
\setlength{\tabcolsep}{3.5pt}
\caption{GelSight Mini on TacQuad with UniTouch (accuracy, \%). The last column is \method\ minus equal fusion.}
\label{tab:tact-unitouch-tacquad-gelsight}
\begin{tabular*}{\linewidth}{@{\extracolsep{\fill}}lrrr@{}}
\toprule
Condition & Equal fusion & \method & \method\ $-$ fusion \\
\midrule
Clean & 48.09 & 60.49 & +12.40 \\
Brightness & 42.74 & 60.77 & +18.03 \\
Contrast & 37.97 & 61.60 & +23.63 \\
Defocus blur & 38.73 & 52.84 & +14.10 \\
Elastic & 37.63 & 47.55 & +9.92 \\
Fog & 38.06 & 52.87 & +14.82 \\
Frost & 33.74 & 48.14 & +14.40 \\
Gaussian noise & 39.85 & 52.68 & +12.83 \\
Glass blur & 47.44 & 64.50 & +17.07 \\
Impulse noise & 45.36 & 53.17 & +7.81 \\
JPEG & 43.31 & 53.45 & +10.14 \\
Motion blur & 39.82 & 56.89 & +17.07 \\
Pixelation & 43.38 & 60.02 & +16.64 \\
Shot noise & 43.50 & 56.48 & +12.98 \\
Snow & 27.61 & 44.49 & +16.87 \\
Zoom blur & 33.02 & 44.56 & +11.54 \\
\bottomrule
\end{tabular*}
\end{table}

%% file: generated/ar_tacquad_digit.tex
\begin{table}[htbp]
\centering\small
\setlength{\tabcolsep}{3.5pt}
\caption{DIGIT on TacQuad with UniTouch (accuracy, \%). The last column is \method\ minus equal fusion.}
\label{tab:tact-unitouch-tacquad-digit}
\begin{tabular*}{\linewidth}{@{\extracolsep{\fill}}lrrr@{}}
\toprule
Condition & Equal fusion & \method & \method\ $-$ fusion \\
\midrule
Clean & 43.46 & 57.81 & +14.35 \\
Brightness & 35.45 & 51.57 & +16.12 \\
Contrast & 31.56 & 63.13 & +31.57 \\
Defocus blur & 38.22 & 54.22 & +16.00 \\
Elastic & 28.29 & 54.46 & +26.17 \\
Fog & 35.03 & 63.26 & +28.23 \\
Frost & 32.30 & 49.24 & +16.94 \\
Gaussian noise & 32.19 & 54.54 & +22.35 \\
Glass blur & 43.32 & 63.97 & +20.65 \\
Impulse noise & 36.24 & 63.51 & +27.28 \\
JPEG & 35.41 & 60.63 & +25.22 \\
Motion blur & 36.95 & 56.41 & +19.46 \\
Pixelation & 42.57 & 58.07 & +15.50 \\
Shot noise & 36.30 & 60.33 & +24.04 \\
Snow & 28.32 & 39.90 & +11.59 \\
Zoom blur & 29.95 & 39.08 & +9.13 \\
\bottomrule
\end{tabular*}
\end{table}

%% file: generated/ar_tacquad_duragel.tex
\begin{table}[htbp]
\centering\small
\setlength{\tabcolsep}{3.5pt}
\caption{DuraGel on TacQuad with UniTouch (accuracy, \%). The last column is \method\ minus equal fusion.}
\label{tab:tact-unitouch-tacquad-duragel}
\begin{tabular*}{\linewidth}{@{\extracolsep{\fill}}lrrr@{}}
\toprule
Condition & Equal fusion & \method & \method\ $-$ fusion \\
\midrule
Clean & 47.79 & 57.47 & +9.68 \\
Brightness & 43.45 & 50.40 & +6.94 \\
Contrast & 45.33 & 53.85 & +8.52 \\
Defocus blur & 52.01 & 54.86 & +2.85 \\
Elastic & 34.38 & 46.55 & +12.17 \\
Fog & 47.39 & 55.66 & +8.27 \\
Frost & 31.57 & 48.94 & +17.37 \\
Gaussian noise & 46.07 & 53.66 & +7.59 \\
Glass blur & 50.52 & 56.14 & +5.62 \\
Impulse noise & 43.12 & 54.28 & +11.16 \\
JPEG & 50.28 & 57.12 & +6.84 \\
Motion blur & 46.87 & 54.67 & +7.79 \\
Pixelation & 42.21 & 53.29 & +11.08 \\
Shot noise & 48.84 & 55.02 & +6.18 \\
Snow & 34.16 & 49.35 & +15.19 \\
Zoom blur & 41.70 & 45.31 & +3.61 \\
\bottomrule
\end{tabular*}
\end{table}

%% file: generated/ar_ofr_splits.tex
\begin{table}[htbp]
\centering\small
\setlength{\tabcolsep}{3.5pt}
\caption{UniTouch on the five ObjectFolder Real test splits (accuracy, \%, objects weighted equally). Default $\beta=0.2$, $g=4$ results; the selected main-table configuration is reported separately. The last column is \method\ minus equal fusion.}
\label{tab:tact-ofr-splits}
\begin{tabular*}{\linewidth}{@{\extracolsep{\fill}}lrrr@{}}
\toprule
Split & Equal fusion & \method & \method\ $-$ fusion \\
\midrule
0 & 41.11 & 59.91 & +18.79 \\
1 & 46.63 & 41.91 & -4.73 \\
2 & 48.71 & 51.93 & +3.22 \\
3 & 38.89 & 57.88 & +18.98 \\
4 & 28.46 & 30.88 & +2.42 \\
\bottomrule
\end{tabular*}
\end{table}

%% file: appendix/unitouch_baselines.tex
\section{Complete comparisons with test-time adaptation baselines}
\label{app:unitouch-baselines}

\subsection{Text configuration and comparison protocol}

All methods use the same checkpoint, ordered inputs, candidate vocabulary,
domain resets and one original prompt per class. Every row shares one
text classifier. Equal fusion normalizes the
fused feature before computing logits, whereas the external methods average
the two branch logits.

The comparison contains ten baselines: nine in Table~\ref{tab:sa-unitouch}
and READ in Appendix~\ref{app:read-results}. The difference tables below carry the six baselines with matched paired
comparison records.
TDA, OGA and StatA operate independently on visual and tactile features and average
their branch predictions; native branch settings, independent state and
pretrained logit scale are retained. TSA and SuMi use their registered
UniTouch baselines. Gradient-based baselines use the declared common
learning rate of $10^{-4}$; \method\ has no encoder optimizer learning
rate.

\paragraph{Parameter-update baselines.}
TENT, SAR, SuMi, DeYO, RoTTA and READ update encoder parameters from the
fused prediction, the average of the image--text and touch--text logits at
the checkpoint scale. TENT minimizes entropy. SAR uses sharpness-aware
entropy minimization, reliable-sample selection, the fixed ImageNet threshold
policy and model recovery. Both update every normalization layer of the
sensory trunks and heads (328 tensors, 0.37M parameters).
SuMi runs its native procedure on the same layers. DeYO keeps
its entropy and PLPD filters and reweighting (patch shuffling with patch
length 4, PLPD threshold 0.2) and updates the normalization layers inside
both trunks (0.20M parameters) with SGD and momentum 0.9. RoTTA keeps its
memory bank and updates the attention projections and normalization layers
of both trunks (309M parameters). READ updates the attention projections
(309M parameters). All use a learning rate of $10^{-4}$, keep
their native number of steps and prediction timing, and reset at the start
of each stream.

\paragraph{Architecture mapping and original defaults.}
We compare READ's and TSA's released defaults with the shared settings
above to clarify the scope of these adapted baselines. Both
methods ship their entry point with learning rate $10^{-4}$ as the
default. READ \citep{yang2024read} declares \texttt{--lr} with default
$10^{-4}$ and optimizes the last cross-modal attention layer with Adam,
betas $(0.9,0.999)$ and no weight decay
(\url{https://github.com/XLearning-SCU/2024-ICLR-READ}); its paper reports
the same initial learning rate of $10^{-4}$. TSA \citep{chen2025tsa}
likewise declares \texttt{--lr} with default $10^{-4}$ and uses Adam with
betas $(0.9,0.999)$ and weight decay $10^{-4}$
(\url{https://github.com/chenmc1996/Uni-Modal-Distribution-Shift}). The
shared learning rate of $10^{-4}$ therefore coincides with the released
default of both methods. The comparison holds this rate fixed under
the shared backbone and batch-size configuration.
The original work uses audio--visual CAV-MAE; our baselines use tactile--visual
backbones. For READ, the branches meet only at fused logits, so its
cross-modal attention update is mapped to attention projections within
the encoders. Tables~\ref{tab:sa-unitouch} and~\ref{tab:sa-tagc}
report TSA; Tables~\ref{tab:read-natural} and~\ref{tab:read-tagc}
report the transferred READ configuration.

The TacQuad comparison includes all 48 sensor--condition streams and
24,576 presentations. The ObjectFolder Real comparison retains all five tests and
3,395 presentations. RCT retains all 9,620 material-recognition inputs.
Object or material groups receive equal weight within a domain, followed
by equal weighting of domains. Presentation-weighted accuracy is recorded in
the source bundles and never substituted for these estimates.

\subsection{READ results on tactile--visual backbones}
\label{app:read-results}

Tables~\ref{tab:read-natural} and~\ref{tab:read-tagc} report the complete
READ results for both backbones under the comparison protocols above.
They evaluate the stated architecture mapping on tactile--visual models;
these results do not describe READ on its original audio--visual backbone.

\input{generated/read_natural}
\input{generated/read_tagc}
\FloatBarrier

\subsection{Changes relative to the initial classifier}

Tables~\ref{tab:unitouch-baselines-selected-same-head-tacquad}
and~\ref{tab:unitouch-baselines-selected-same-head-natural} compare each
reported external method with its single-prompt frozen reference. The initial classifier is
shared within these comparisons; native prediction timing and the StatA
precision qualification remain as specified above. The tables retain
every reported paired difference.

\input{generated/unitouch_baselines_selected_same_head_tacquad}
\input{generated/unitouch_baselines_selected_same_head_natural}
\FloatBarrier

\subsection{Numerical precision and prediction validity}

StatA's native FP32 computation produced nonfinite outputs on 52 TacQuad
and 217 RCT batches. A uniform FP64 implementation of the same equations and settings
completed the full streams without this failure. The comparison reports
that implementation as StatA, with FP64 specified in the table captions.
It changes 10,438 predictions among 23,744 finite TacQuad rows,
470 among 6,160 finite RCT rows, and 399 of the 3,395 ObjectFolder
Real predictions. Precision therefore affects more than the
previously nonfinite rows. The FP64 score is a result for this numerical
baseline, rather than a bit-equivalent reproduction of the native FP32 run;
the latter has no valid full-stream TacQuad or RCT score.
Table~\ref{tab:unitouch-baselines-stata-precision} gives the coverage of
both implementations.

\input{generated/unitouch_baselines_stata_precision}
\FloatBarrier

%% file: generated/read_natural.tex
\begin{table}[htbp]
\centering
\footnotesize
\caption{READ natural-shift accuracy (\%) on the tactile--visual backbones. The four panels, input order and aggregation match Table~\ref{tab:sa-unitouch}; each Mean weights its four panels equally.}
\label{tab:read-natural}
\setlength{\tabcolsep}{3pt}
\begin{tabular}{@{}lrrrrr@{\hspace{8pt}}rrrrr@{}}
\toprule
 & \multicolumn{5}{c}{UniTouch} & \multicolumn{5}{c}{AnyTouch} \\
\cmidrule(lr){2-6}\cmidrule(l){7-11}
Method & \shortstack{TacQuad\\clean} & \shortstack{TacQuad\\corr.} & OFR & RCT & Mean & \shortstack{TacQuad\\clean} & \shortstack{TacQuad\\corr.} & OFR & RCT & Mean \\
\midrule
READ & 9.9 & 9.1 & 17.9 & 26.4 & 15.9 & 12.1 & 9.7 & 18.0 & 3.3 & 10.8 \\
\bottomrule
\end{tabular}
\end{table}

%% file: generated/read_tagc.tex
\begin{table}[htbp]
\centering
\fontsize{8.5}{10}\selectfont
\caption{READ TAG-C accuracy (\%) at severity 3 on the tactile--visual backbones. Scenarios and aggregation match Table~\ref{tab:sa-tagc}; each model Mean weights three scenarios equally, and Overall weights all six.}
\label{tab:read-tagc}
\setlength{\tabcolsep}{3pt}
\begin{tabular}{@{}lccccccccc@{}}
\toprule
& \multicolumn{4}{c}{AnyTouch} & \multicolumn{4}{c}{UniTouch} & Overall \\
\cmidrule(lr){2-5}\cmidrule(lr){6-9}\cmidrule(l){10-10}
Method & Visual & Tactile & Mixed & Mean & Visual & Tactile & Mixed & Mean & Mean \\
\midrule
READ & 6.4 & 10.2 & 0.1 & 5.6 & 1.3 & 23.2 & 6.5 & 10.3 & 7.9 \\
\bottomrule
\end{tabular}
\end{table}

%% file: generated/unitouch_baselines_selected_same_head_tacquad.tex
\begin{table}[htbp]
\centering\small
\setlength{\tabcolsep}{3.5pt}
\caption{Reported external method minus equal fusion with one prompt per class on TacQuad. Differences are in percentage points. StatA uses FP64 arithmetic.}
\label{tab:unitouch-baselines-selected-same-head-tacquad}
\begin{tabular*}{\linewidth}{@{\extracolsep{\fill}}lrr@{}}
\toprule
Method & Clean & Corruptions \\
\midrule
READ & -36.50 & -30.23 \\
SuMi & -0.08 & -0.53 \\
TDA & +2.68 & +4.20 \\
OGA & +5.65 & +6.71 \\
StatA & +4.86 & +5.82 \\
TSA & -2.98 & -3.08 \\
\bottomrule
\end{tabular*}
\par\smallskip\begin{minipage}{\linewidth}\footnotesize
Methods start from the same single-prompt classifier and follow their registered output timing.
\end{minipage}
\end{table}

%% file: generated/unitouch_baselines_selected_same_head_natural.tex
\begin{table}[htbp]
\centering\small
\setlength{\tabcolsep}{3.5pt}
\caption{Reported external method minus equal fusion with one prompt per class on natural shifts. Differences are in percentage points. StatA uses FP64 arithmetic.}
\label{tab:unitouch-baselines-selected-same-head-natural}
\begin{tabular*}{\linewidth}{@{\extracolsep{\fill}}lrr@{}}
\toprule
Method & \shortstack[l]{ObjectFolder\\Real} & RCT \\
\midrule
READ & -22.81 & -21.65 \\
SuMi & -1.42 & -21.85 \\
TDA & +1.82 & -10.90 \\
OGA & +7.31 & -39.02 \\
StatA & +5.48 & +12.15 \\
TSA & -3.21 & -36.18 \\
\bottomrule
\end{tabular*}
\par\smallskip\begin{minipage}{\linewidth}\footnotesize
Methods start from the same single-prompt classifier and follow their registered output timing.
\end{minipage}
\end{table}

%% file: generated/unitouch_baselines_stata_precision.tex
\begin{table}[htbp]
\centering\small
\setlength{\tabcolsep}{3.5pt}
\caption{StatA numerical precision diagnostic on the completed full streams.}
\label{tab:unitouch-baselines-stata-precision}
\begin{tabular*}{\linewidth}{@{\extracolsep{\fill}}lrrrr@{}}
\toprule
Dataset & FP32 invalid batches & Finite rows & Disagreements & FP32 accuracy \\
\midrule
TacQuad & 52 & 23744 & 10438 & --- \\
\shortstack[l]{ObjectFolder\\Real} & 0 & 3395 & 399 & 42.00 \\
RCT & 217 & 6160 & 470 & --- \\
\bottomrule
\end{tabular*}
\par\smallskip\begin{minipage}{\linewidth}\footnotesize
Invalid batches contain nonfinite FP32 outputs; --- denotes undefined full-stream accuracy. Accuracy is macro accuracy (\%). Disagreements count changed predictions among rows in entirely finite native FP32 batches. FP64 evaluates the same formulas and settings uniformly. FP64 changes predictions even where FP32 is finite, so it is reported as a numerical baseline rather than a bit-equivalent reproduction.
\end{minipage}
\end{table}

%% file: appendix/unitouch_tagc.tex
\section{TAG-C corruption robustness with UniTouch}
\label{app:unitouch-tagc}

\subsection{Data streams and adaptation configurations}

The UniTouch extension retains all 4,126 original TAG-C frame identities.
Fifteen visual corruption conditions contribute 61,890 presentations;
seven tactile conditions contribute 28,882. All use the original
severity-three assets and the same 20 material categories, including
\textit{Others}. The uncorrupted modality remains available in each
direction. These two streams are distinct from the 94,898-presentation
mixed stream evaluated with both backbones, which additionally includes simultaneous
corruptions and clean observations.

Each direction starts with fresh adaptation state. Its original row order
is retained without resets between corruption conditions. External batches
contain 16 observations, with the final two observations evaluated as their
own batch. No observations are dropped or joined to an earlier batch.

Equal fusion and the baselines use one original prompt per class.
\method\ uses the same prompts and predicts a complete batch before incorporating it
into the shared memory. Anchor Realignment uses the fixed
64-step classification refinement with a 16-observation warm-up, while
all encoders remain frozen. The default control uses $\beta=0.2$, $g=4$;
the main-table Visual result uses the selected $\beta=0.4$, $g=8$ setting.
Tactile retains the defaults. Labels are used by the separate scoring
and panel-level parameter selection, not during adaptation.

READ, TDA, OGA and TSA use the single-prompt classifier and the registered
serving conventions described in Appendix~\ref{app:unitouch-baselines}.
They retain every observation in both directions. TDA and OGA keep
independent visual and tactile adaptation state and combine branch outputs
equally. READ updates the selected attention projections in the visual,
tactile and text branches. TSA retains its native post-update traversal sum
and the declared shared-learning-rate override. We report recorded optimizer
calls without assuming one update per batch.

StatA's native $k=3$ operation is undefined for each original final batch
of two observations. Both the FP64 numerical implementation and the FP32
sensitivity run retain those four undefined observations in their validity
masks. Table~\ref{tab:sa-tagc} reports the FP64 baseline over the remaining
90,768 observations. We do not impute predictions for the four, join the
tail to a preceding batch, or drop any other observation. The FP32
sensitivity run gives 52.50 and 53.58 on the two directions, against 52.59
and 53.69 for FP64.

Touch and Go is included in UniTouch's documented pretraining sources.
These results measure corruption robustness on a documented pretraining
source, complementing the primary natural-shift streams.

\subsection{Aggregation}

Accuracy is the fraction of correct presentations within each corruption.
Each directional score averages its corruption conditions equally, and the
UniTouch directional overall weights the visual and tactile directions
equally. Thus the 15 visual conditions do not receive more total weight than
the seven tactile conditions. In the main table, each backbone mean equally
weights Visual, Tactile, and Mixed, and Overall equally weights all six
scenarios. This presentation-based calculation differs from the group-balanced
metric of the three primary datasets.

Table~\ref{tab:sa-tagc} compares \method\ with the baselines and
single-prompt equal fusion.
Table~\ref{tab:unitouch-tagc-effects} retains per-corruption differences
relative to single-prompt equal fusion at the default $\beta=0.2$, $g=4$
settings. Its visual mean differs from the selected main-table result.

\input{generated/ar_unitouch_tagc_effects}
\subsection{Implementation and numerical status}

The following records specify the adaptation procedures and numerical
status associated with the reported table entries.

READ updates attention projections in the visual, tactile and text branches
and returns predictions after adaptation, using its original loss and the
shared learning rate of $10^{-4}$. Table~\ref{tab:read-tagc} reports its
results in both directions and on Mixed. Appendix~\ref{app:unitouch-baselines}
specifies the architecture mapping and the parameters receiving the loss.

SuMi retains feature smoothing, its time schedule and entropy selection,
and updates LayerNorm affine parameters in all three branches. The visual
and Mixed entries retain their completed native mixed-precision runs.
For the tactile entry, native runs encountered nonfinite logits; correcting
the reciprocal gradient scale alone subsequently left the scale at zero
after 7,136 observations. We therefore reran the complete tactile stream
from the original checkpoint in FP32, disabling autocast and gradient
scaling while retaining the original loss, thresholds, learning rate,
parameter groups and post-update predictions. This precision recovery
completed all 28,882 observations with 1,806 optimizer calls,
finite predictions and finite gradients at every executed update. Every
batch selected by the native rule triggered an optimizer call; parameters
outside the declared groups remained unchanged. Independent recomputation
gives 45.38\% tactile accuracy. The failed mixed-precision
trajectories contribute no accuracy to the table.

\subsection{Complete matched Mixed comparison}
\label{app:unitouch-mixed}

The UniTouch Mixed extension uses the same 94,898 ordered presentations,
20 classes and 117 capture directories as AnyTouch. The stream contains
4,126 clean, 30,945 visually corrupted, 14,441 tactile-corrupted and
45,386 jointly corrupted presentations. Each method starts with fresh
state and then runs continuously, without resets at corruption boundaries
apart from its native recovery rule. All methods use 5,930 batches of 16
followed by batches of 14 and 4, retaining every presentation. Parameter
updates use the shared $10^{-4}$ learning rate and the previously declared
parameter groups, native losses, thresholds and serving timing. TDA, OGA
and StatA maintain separate modality states and average their outputs.
The default SenseAlign control uses $g=4$, $\beta=0.2$,
the 16-observation warm-up and 64-step refinement, with one prompt per
class, and scores 55.10\%. The main-table Mixed result uses the selected
$\beta=0.4$, $g=8$ configuration and scores 57.13\%
(Appendix~\ref{app:hyperparameters}); the remaining settings are unchanged.

The independent scorer verifies full predictions, input identity and order,
class and prompt order, checkpoint, code and configuration hashes, and
all batch records. Direct image-to-output frozen inference agrees with
all 94,898 cached frozen predictions (47,042 correct). The complete
baseline counts and selected SenseAlign predictions supply the Mixed
entries in Table~\ref{tab:sa-tagc}.
Both backbone means equally weight their three scenarios, and Overall
weights all six, using full-precision constituent scores before rounding.
The comparison describes method performance on this recorded trajectory
under the specified checkpoint, inputs and adaptation configurations.

The UniTouch Mixed StatA row retains the previously accepted FP64 numerical
implementation. Its FP32 counterpart is finite in all 5,932 batches but
has different predicted classes on 281 presentations. The table reports
the FP64 configuration throughout. The Mixed tail
sizes support its native $k=3$ operation; the four undefined observations
in the original directional runs remain excluded only from those rows.

The unmodified Mixed SuMi run completed with 15,187 correct predictions
and 4,264 optimizer calls; its AMP scale remained positive. This separate
result occupies the Mixed cell; the tactile-directional entry uses the FP32 recovery described above.

%% file: generated/ar_unitouch_tagc_effects.tex
\begin{table}[htbp]
\centering\footnotesize
\setlength{\tabcolsep}{2pt}
\caption{UniTouch TAG-C differences from single-prompt equal fusion at default $\beta=0.2$, $g=4$ (percentage points).}
\label{tab:unitouch-tagc-effects}
\begin{tabular*}{\linewidth}{@{\extracolsep{\fill}}lr@{}}
\toprule
Condition & \method\ minus fusion \\
\midrule
Visual mean & +4.04 \\
Tactile mean & +10.53 \\
Overall mean & +7.29 \\
\midrule
\multicolumn{2}{@{}l}{\textit{Visual corruptions}} \\
Brightness & +4.90 \\
Defocus blur & +5.40 \\
Fog & +5.62 \\
Gaussian noise & +1.19 \\
Impulse noise & +3.56 \\
Motion blur & +2.69 \\
Shot noise & +4.19 \\
Contrast & +3.76 \\
Elastic & +4.99 \\
Frost & +4.58 \\
Glass blur & +3.01 \\
JPEG & +3.95 \\
Pixelation & +4.14 \\
Snow & +4.48 \\
Zoom blur & +4.12 \\
\midrule
\multicolumn{2}{@{}l}{\textit{Tactile corruptions}} \\
Brightness & +7.59 \\
Defocus blur & +6.62 \\
Gaussian noise & +14.13 \\
Impulse noise & +11.05 \\
Motion blur & +10.59 \\
Contrast & +17.72 \\
Elastic & +6.03 \\
\bottomrule
\end{tabular*}
\end{table}

%% file: appendix/matched_readout_controls.tex
\section{Matched comparisons of readout rules}
\label{app:matched-readouts}

These default-setting comparisons retain $g=4$, including AnyTouch OFR,
whose selected main-table setting is $g=1$. They ask whether the semantic weighting rule adds value
after holding feature realignment fixed. Each corrected serving rule uses
the same encoders, class vocabulary, single-prompt head, batch order,
memory writes and fitted maps. Only their serving weights change; frozen references and the original
Realign-only control are listed separately below.
All eleven primary panels are retained: four UniTouch natural-shift
panels, AnyTouch OFR in released order, and six TAG-C scenarios.

\paragraph{Weighting rules.}
Let $q^v$ and $q^t$ denote the realigned branch logits in
Equation~\ref{eq:prediction}. Set $p^m=\operatorname{softmax}(q^m)$,
$c_m=\max_j p^m_j$, and $H_m=-\sum_j p^m_j\log p^m_j$.
Maximum-probability weighting uses $w_v=c_v/(c_v+c_t)$; inverse-entropy
weighting uses $w_v=(H_t+\epsilon)/(H_v+H_t+2\epsilon)$, with
$\epsilon=10^{-12}$. These rules use the current input's branch logits
at the original checkpoint scale, without a temperature search.
Maximum-probability selection assigns all weight to the more confident
branch. Alignment selection instead chooses the branch with larger
historical Trust weight. Ties within $10^{-12}$ use equal weights,
including Trust warm-up. The prefix control follows Trust until the first
completed batch containing at least 256 cumulative inputs, then keeps
the next-batch weight fixed while realignment continues to update.

Corrected single-branch readouts use $w_v=1$ or $0$ with the same
multimodally fitted map. They are readout controls, not independently
adapted unimodal methods. Frozen branches and frozen equal fusion are
also reported. Corrected equal weights and Realign only coincide
on every panel.

\paragraph{Results and uncertainty.}
Table~\ref{tab:readout-summary} preserves the three original aggregate
definitions rather than combining them into a new overall measure.
The TAG-C mean is 60.44\% for SenseAlign, 59.23\% for corrected equal
weights, 59.51\% for maximum-probability weights and 59.64\% for
inverse-entropy weights. The UniTouch natural mean also favors
SenseAlign over the probability and entropy rules. On AnyTouch OFR,
those two rules and corrected equal weights have higher point estimates.
Tables~\ref{tab:readout-natural} and~\ref{tab:readout-tagc} retain every
primary panel and comparator.

The intervals in Table~\ref{tab:readout-intervals} use 5,000 paired
bootstrap draws at the original grouping level: named targets for
TacQuad, objects for OFR, materials for RCT, and capture directories for
TAG-C. Repeated groups share the same draw across conditions and OFR
splits; TAG-C also shares capture draws across directions and backbones.
Each draw retains the original within-panel aggregation, followed by
the specified equal-panel mean. These are intervals conditional on the
saved adaptation histories: maps and weights are not refitted after
resampling. They do not correct for multiple comparisons or development
selection and do not measure independent test-set generalization.

Across the six TAG-C scenarios, SenseAlign minus corrected equal weights
is $+1.22$ points with interval $[+0.89,+1.56]$; against probability and
inverse-entropy weights the differences are $+0.94$ and $+0.80$ points.

\input{generated/sa_readout_summary}
\input{generated/sa_readout_intervals}
\input{generated/sa_readout_natural}
\input{generated/sa_readout_tagc}

\paragraph{Matching history and class grouping.}
The additional comparison in Table~\ref{tab:history-margin} covers
UniTouch TacQuad clean, TacQuad corrupted and RCT, and the six TAG-C panels.
It preserves Trust's accepted observations, shared pseudo-label groups,
sample weights, class means and prediction-before-update timing.
Let $z_{c,b,(1)}^m$ and $z_{c,b,(2)}^m$ be the two largest entries of the
standardized score vector in Equation~\ref{eq:standardized_scores}.
History-grouped margin replaces assigned-class support with the top-two gap:
\begin{equation}
 D_b^m=\frac{1}{|\mathcal C_b|}\sum_{c\in\mathcal C_b}
 \big(z_{c,b,(1)}^m-z_{c,b,(2)}^m\big),\qquad
 [w_b^v,w_b^t]=\operatorname{softmax}\big(4[D_b^v,D_b^t]\big).
 \label{eq:history-margin}
\end{equation}
The current-sample margin control applies the same standardization and
top-two gap to each input's raw branch scores, using the same gain and
readiness condition. All three rules combine identical corrected branch
scores. Equal gain does not imply identical weight sharpness across signals.
Intervals use the paired, fixed-history bootstrap described above.

\input{generated/sa_history_margin}

\paragraph{Interpreting the UniTouch exceptions.}
The two exceptions are consistent with Trust assigning too little weight
to vision. On UniTouch Tactile and Mixed, history-grouped margin raises
the mean visual weight from 0.709 to 0.755 and from 0.610 to 0.658,
respectively. Its net gains arise on predictions receiving more visual
weight; both panels have a more accurate corrected visual branch
(Table~\ref{tab:readout-tagc}). Trust measures support for shared predicted
classes, which need not calibrate the relative usefulness of the corrected
branches. Its more balanced weights can therefore retain too much tactile
influence in these streams. Both fusion rules nevertheless outperform
corrected vision alone, supporting a stronger visual contribution while
retaining complementary touch. This interpretation describes the saved
weight and prediction differences; it does not establish pseudo-label
errors as their underlying cause.

\paragraph{Frozen reference provenance.}
The paired controls use frozen predictions from the same method replay.
On UniTouch Mixed, this frozen reference differs from the separately
executed main-table baseline on one of 94,898 predictions, with 47,041
rather than 47,042 correct. Both round to 49.57\%; the six-scenario mean
changes by less than 0.0002 points. The difference does not enter any
contrast between corrected weighting rules. All SenseAlign predictions
in these controls match their archived method outputs.

%% file: generated/sa_readout_summary.tex
\begin{table}[htbp]
\centering\footnotesize
\caption{Matched readout comparisons (accuracy, \%). All use one prompt per class. The natural mean weights the four UniTouch panels equally; OFR uses AnyTouch in the five splits in released order; TAG-C weights its six model--scenario combinations equally. Corrected rules share the same realignment fits.}
\label{tab:readout-summary}
\setlength{\tabcolsep}{3pt}
\begin{adjustbox}{max width=\linewidth}
\begin{tabular}{@{}lrrr@{}}
\toprule
Readout & UniTouch natural & AnyTouch OFR & TAG-C mean \\
\midrule
Frozen equal fusion & 43.67 & 71.67 & 53.85 \\
Frozen vision & 54.93 & 37.31 & 52.50 \\
Frozen touch & 7.20 & 74.62 & 41.94 \\
Realign only & 46.44 & 78.27 & 59.23 \\
Corrected equal weights & 46.44 & 78.27 & 59.23 \\
Maximum-probability weights & 49.07 & 78.31 & 59.51 \\
Inverse-entropy weights & 50.53 & 78.28 & 59.64 \\
Maximum-probability selection & 50.73 & 77.04 & 58.58 \\
Alignment selection & 56.70 & 75.15 & 54.19 \\
Weights fixed after 256 inputs & 44.37 & 76.58 & 60.21 \\
Corrected vision readout & 56.40 & 53.71 & 57.97 \\
Corrected touch readout & 11.49 & 77.07 & 46.69 \\
SenseAlign & 56.32 & 76.08 & 60.44 \\
\bottomrule
\end{tabular}
\end{adjustbox}
\end{table}

%% file: generated/sa_readout_intervals.tex
\begin{table}[htbp]
\centering\footnotesize
\caption{SenseAlign minus each readout in Table~\ref{tab:readout-summary} (percentage points; paired 95\% intervals). Intervals condition on fixed adaptation histories and use 5,000 group bootstrap draws. They do not include uncertainty from refitting or parameter selection.}
\label{tab:readout-intervals}
\setlength{\tabcolsep}{3pt}
\begin{adjustbox}{max width=\linewidth}
\begin{tabular}{@{}lrrr@{}}
\toprule
Readout & UniTouch natural & AnyTouch OFR & TAG-C mean \\
\midrule
Frozen equal fusion & $+12.65\;[+8.05,+17.36]$ & $+4.40\;[-4.29,+13.20]$ & $+6.59\;[+5.13,+7.99]$ \\
Frozen vision & $+1.39\;[-0.96,+3.94]$ & $+38.77\;[+25.49,+51.43]$ & $+7.94\;[+6.03,+9.85]$ \\
Frozen touch & $+49.12\;[+39.99,+58.03]$ & $+1.46\;[-3.09,+5.75]$ & $+18.50\;[+15.35,+21.09]$ \\
Realign only & $+9.87\;[+6.52,+13.30]$ & $-2.19\;[-9.32,+5.00]$ & $+1.22\;[+0.89,+1.56]$ \\
Corrected equal weights & $+9.87\;[+6.52,+13.30]$ & $-2.19\;[-9.32,+5.00]$ & $+1.22\;[+0.89,+1.56]$ \\
Maximum-probability weights & $+7.25\;[+4.60,+9.97]$ & $-2.24\;[-9.31,+4.91]$ & $+0.94\;[+0.61,+1.25]$ \\
Inverse-entropy weights & $+5.78\;[+3.66,+7.91]$ & $-2.20\;[-9.36,+4.71]$ & $+0.80\;[+0.41,+1.17]$ \\
Maximum-probability selection & $+5.59\;[+3.81,+7.49]$ & $-0.96\;[-7.79,+5.72]$ & $+1.86\;[+1.37,+2.35]$ \\
Alignment selection & $-0.39\;[-0.95,+0.08]$ & $+0.92\;[-1.09,+2.63]$ & $+6.25\;[+5.19,+7.31]$ \\
Weights fixed after 256 inputs & $+11.95\;[+9.63,+14.17]$ & $-0.51\;[-1.26,+0.13]$ & $+0.23\;[+0.10,+0.36]$ \\
Corrected vision readout & $-0.08\;[-0.96,+0.94]$ & $+22.37\;[+10.26,+34.08]$ & $+2.47\;[+1.45,+3.55]$ \\
Corrected touch readout & $+44.82\;[+35.27,+54.04]$ & $-0.99\;[-4.45,+1.95]$ & $+13.75\;[+12.10,+15.21]$ \\
\bottomrule
\end{tabular}
\end{adjustbox}
\end{table}

%% file: generated/sa_readout_natural.tex
\begin{table}[htbp]
\centering\footnotesize
\caption{All five natural-shift panels for the matched readout comparison (accuracy, \%). TacQuad clean and corrupted, OFR and RCT use UniTouch; the final column uses AnyTouch OFR. Grouping, histories and prompts are fixed within each panel.}
\label{tab:readout-natural}
\setlength{\tabcolsep}{3pt}
\begin{adjustbox}{max width=\linewidth}
\begin{tabular}{@{}lrrrrr@{}}
\toprule
Readout & \shortstack{TacQuad\\clean} & \shortstack{TacQuad\\corrupted} & \shortstack{UniTouch\\OFR} & RCT & \shortstack{AnyTouch\\OFR} \\
\midrule
Frozen equal fusion & 46.45 & 39.38 & 40.76 & 48.09 & 71.67 \\
Frozen vision & 54.47 & 53.73 & 48.70 & 62.81 & 37.31 \\
Frozen touch & 0.55 & 0.55 & 16.24 & 11.45 & 74.62 \\
Realign only & 49.96 & 44.71 & 43.10 & 47.99 & 78.27 \\
Corrected equal weights & 49.96 & 44.71 & 43.10 & 47.99 & 78.27 \\
Maximum-probability weights & 53.24 & 49.28 & 45.09 & 48.65 & 78.31 \\
Inverse-entropy weights & 54.67 & 51.25 & 47.92 & 48.29 & 78.28 \\
Maximum-probability selection & 56.23 & 52.52 & 46.69 & 47.48 & 77.04 \\
Alignment selection & 58.59 & 54.20 & 49.91 & 64.11 & 75.15 \\
Weights fixed after 256 inputs & 58.59 & 54.02 & 48.59 & 16.28 & 76.58 \\
Corrected vision readout & 57.16 & 54.58 & 50.00 & 63.84 & 53.71 \\
Corrected touch readout & 4.94 & 5.60 & 19.74 & 15.69 & 77.07 \\
SenseAlign & 58.59 & 54.03 & 48.50 & 64.14 & 76.08 \\
\bottomrule
\end{tabular}
\end{adjustbox}
\end{table}

%% file: generated/sa_readout_tagc.tex
\begin{table}[htbp]
\centering\footnotesize
\caption{All six single-prompt TAG-C scenarios for the matched readout comparison (accuracy, \%). A and U denote AnyTouch and UniTouch; V, T and M denote visual corruption, tactile corruption and Mixed. Corrected equal weights and Realign only coincide.}
\label{tab:readout-tagc}
\setlength{\tabcolsep}{3pt}
\begin{adjustbox}{max width=\linewidth}
\begin{tabular}{@{}lrrrrrr@{}}
\toprule
Readout & A--V & A--T & A--M & U--V & U--T & U--M \\
\midrule
Frozen equal fusion & 56.08 & 59.13 & 55.17 & 52.85 & 50.28 & 49.57 \\
Frozen vision & 49.56 & 62.09 & 51.99 & 47.37 & 55.14 & 48.85 \\
Frozen touch & 47.60 & 39.42 & 42.41 & 46.61 & 35.81 & 39.78 \\
Realign only & 63.03 & 66.52 & 62.73 & 56.05 & 54.25 & 52.77 \\
Corrected equal weights & 63.03 & 66.52 & 62.73 & 56.05 & 54.25 & 52.77 \\
Maximum-probability weights & 63.08 & 66.77 & 62.87 & 55.84 & 55.29 & 53.18 \\
Inverse-entropy weights & 62.97 & 66.94 & 62.84 & 55.59 & 56.03 & 53.47 \\
Maximum-probability selection & 61.81 & 66.28 & 61.56 & 54.29 & 55.13 & 52.39 \\
Alignment selection & 56.60 & 54.73 & 50.94 & 50.78 & 59.97 & 52.14 \\
Weights fixed after 256 inputs & 63.08 & 65.87 & 62.29 & 56.81 & 60.56 & 52.65 \\
Corrected vision readout & 56.57 & 69.09 & 59.26 & 50.77 & 59.96 & 52.17 \\
Corrected touch readout & 56.57 & 47.80 & 50.94 & 47.91 & 36.32 & 40.57 \\
SenseAlign & 62.41 & 65.99 & 61.45 & 56.88 & 60.81 & 55.10 \\
\bottomrule
\end{tabular}
\end{adjustbox}
\end{table}

%% file: generated/sa_history_margin.tex
\begin{table}[htbp]
\centering\footnotesize
\caption{Matched-history reliability comparison (accuracy, \%).
The final column gives Trust minus history-grouped margin in percentage points,
with paired 95\% intervals. Means weight the three TacQuad/RCT or six TAG-C panels equally.}
\label{tab:history-margin}
\setlength{\tabcolsep}{3pt}
\begin{adjustbox}{max width=\linewidth}
\begin{tabular}{@{}llrrrr@{}}
\toprule
Model & Scenario & Sample margin & History margin & Trust & Difference [95\% interval] \\
\midrule
UniTouch & TacQuad clean & 52.66 & 57.51 & 58.59 & $+1.08$ [$ +0.22, +2.28 $] \\
UniTouch & TacQuad corrupted & 49.19 & 52.91 & 54.03 & $+1.12$ [$ +0.27, +2.28 $] \\
UniTouch & RCT & 47.46 & 56.37 & 64.14 & $+7.78$ [$ +2.36, +13.43 $] \\
\midrule
AnyTouch & TAG-C Visual & 60.62 & 60.68 & 62.41 & $+1.74$ [$ +1.18, +2.29 $] \\
AnyTouch & TAG-C Tactile & 64.10 & 62.26 & 65.99 & $+3.73$ [$ +3.17, +4.27 $] \\
AnyTouch & TAG-C Mixed & 60.13 & 58.02 & 61.45 & $+3.43$ [$ +2.90, +3.92 $] \\
\midrule
UniTouch & TAG-C Visual & 54.82 & 56.81 & 56.88 & $+0.07$ [$ -0.23, +0.37 $] \\
UniTouch & TAG-C Tactile & 56.22 & 61.57 & 60.81 & $-0.76$ [$ -1.03, -0.48 $] \\
UniTouch & TAG-C Mixed & 53.14 & 55.75 & 55.10 & $-0.65$ [$ -0.91, -0.39 $] \\
\midrule
UniTouch & TacQuad/RCT mean & 49.77 & 55.60 & 58.92 & $+3.33$ [$ +1.43, +5.30 $] \\
Both & TAG-C mean & 58.17 & 59.18 & 60.44 & $+1.26$ [$ +1.00, +1.52 $] \\
\bottomrule
\end{tabular}
\end{adjustbox}
\end{table}

%% file: appendix/realignment_diagnostic.tex
\section{Retrospective analysis of Anchor Realignment}
\label{app:realignment-diagnostic}
\input{generated/realignment_values}

We examine how the fitting stages change feature geometry and class scores
with AnyTouch on ObjectFolder Real split 0, using 704 observations and
seven classes. We reconstruct the final shared
memory, the Procrustes initialization, and classification refinement with the
original solver. Encoders and semantic anchors remain fixed. Both branches
receive the same correction and are fused with weights of one half, so
Alignment Trust does not contribute to this diagnostic. All observations
are retained and weighted equally in the reported means and accuracies.
True labels are used only for analysis; the fits use the shared-memory
pseudo-labels. Each final memory includes the replayed stream, and the
results describe retrospective application to those same observations.
The subsequent factor and normalization controls analyze these fitted maps.
In the compact figure and table labels, ``Raw'' means equal-weight fusion,
``Procrustes'' means realignment without refinement, and ``CE'' means
feature realignment after the cross-entropy classification refinement of
Section~\ref{sec:realign}.

\subsection{Geometric matching and class discrimination}

For each stage, we compute diagnostic cosine scores from the normalized
fusion of the corrected unit branches,
\begin{equation}
 z_i=\nu\big((\phi(u_i^v)+\phi(u_i^t))/2\big),\qquad
 r_{ic}=z_i a_c^\top.
 \label{eq:ar-diagnostic-score}
\end{equation}
The raw stage uses $\phi=\nu$. These scores omit the checkpoint logit
scale and normalize the fused feature, unlike the serving logits in
Eq.~\ref{eq:prediction}. This additional positive scale leaves predictions
unchanged but changes score magnitudes. We decompose each score and define
the gap to the strongest incorrect class as
\begin{equation}
 \begin{aligned}
 r_{i,\mathrm{avg}}&=\frac1C\sum_j r_{ij},\qquad
 r_{ic}=r_{i,\mathrm{avg}}+(r_{ic}-r_{i,\mathrm{avg}}),\\
 \Delta_i&=r_{iy_i}-\max_{j\ne y_i}r_{ij}.
 \end{aligned}
 \label{eq:ar-score-decomposition}
\end{equation}
The first term is identical for all classes of an observation. It cancels
in score differences and softmax probabilities. The gap $\Delta_i$ uses
the true label $y_i$ and is an evaluation diagnostic, not an inference-time
selection or weighting signal.

\begin{figure}[!ht]
 \centering
 \includegraphics[width=\linewidth]{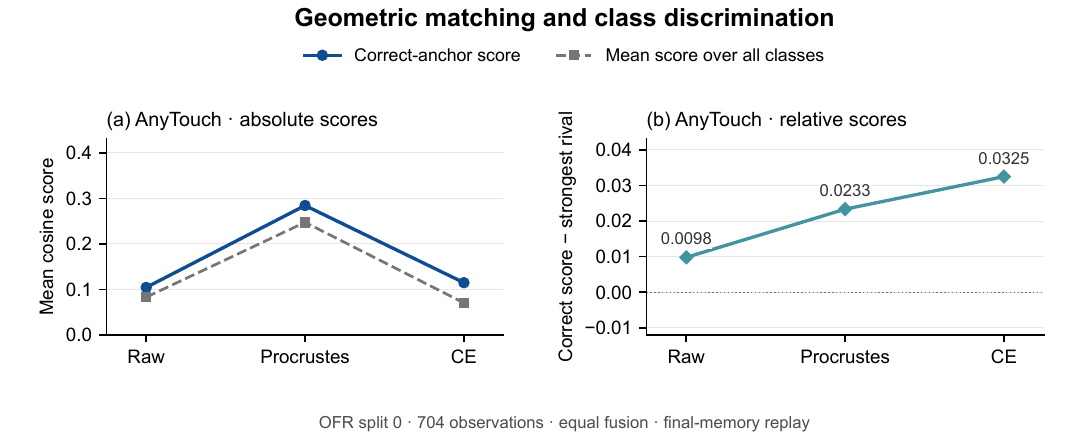}
 \caption{\textbf{Geometric matching and class discrimination.}
Raw denotes uncorrected features; Procrustes and CE denote geometric
initialization and subsequent cross-entropy refinement.
(a) Mean cosine similarity to the correct-class text embedding and the
mean across all class embeddings. (b) Mean gap between the correct-class
score and the strongest incorrect-class score.
Scores are evaluated retrospectively using the final memory.}
 \label{fig:ar-score-stages}
\end{figure}

Procrustes increases mean correct-anchor similarity, but an increase in
the mean score over all classes accounts for \arAnyCommonGain{} of that
increase
(Figure~\ref{fig:ar-score-stages}). This is a ratio of changes in dataset
means. Accuracy is computed separately from predicted classes.
Classification refinement then reduces the mean
correct-anchor score, while reducing the mean
strongest-rival score more. The mean gap therefore increases
(Table~\ref{tab:ar-stage-diagnostic}). These quantities summarize
the observation population.

\begin{table}[!ht]
 \centering\small
 \caption{Stage scores and retrospective accuracy on the same ObjectFolder Real replay.
 Correct and Rival are the means of $r_{iy_i}$ and
 $\max_{j\ne y_i}r_{ij}$; Gap is their difference.
 The last row applies the same refined map with $\beta=1$
 without refitting a map optimized at $\beta=0.2$.}
 \label{tab:ar-stage-diagnostic}
 \input{generated/realignment_stages}
\end{table}

\subsection{The component shared by the semantic anchors}

Let $a_{\mathrm{avg}}=C^{-1}\sum_c a_c$ denote the average across classifier
anchors, distinct from an individual class anchor $a_c$.
The following identity separates its squared norm from variation across
classes,
\begin{equation}
 \frac1C\sum_c\|a_c\|^2
 =\|a_{\mathrm{avg}}\|^2+
   \frac1C\sum_c\|a_c-a_{\mathrm{avg}}\|^2.
 \label{eq:ar-anchor-energy}
\end{equation}
The ratio of the first term to the left-hand side is \arAnyEnergy{} for
AnyTouch. Here ``energy'' means squared
vector norm. Classification depends on the differences between anchor
scores. For any feature $z$, the score term $z a_{\mathrm{avg}}^\top$ is
shared across classes. Moving toward this common component can raise
absolute anchor similarities without improving class ranking. The
decomposition measures the shared component of the evaluated anchors;
the remaining class-specific component determines their score differences.

\subsection{Geometry of the residual map}

For unit $u$ and orthogonal $R$, the serving transform simplifies to
$\phi(u)=\nu(uB)$, where $B=(1-\beta)I+\beta R$. Although $R$ preserves
inner products, generally
\begin{equation}
 BB^\top=\big((1-\beta)^2+\beta^2\big)I
          +\beta(1-\beta)(R+R^\top)\ne I.
 \label{eq:ar-residual-geometry}
\end{equation}
Thus residual mixing introduces directional scaling as well as a change
of orientation. At $\beta=0.2$, the triangle and reverse-triangle
inequalities give $0.6\|u\|\le\|uB\|\le\|u\|$, so the singular values of
$B$ lie in $[0.6,1]$. The map is invertible; for fixed memory, its direction
map has inverse $z\mapsto\nu(zB^{-1})$. This invertible transform adapts the
readout while retaining the information in each individual feature.

Writing $\theta=\angle(u,uR)$ gives
$\tan\angle(u,uB)=\beta\sin\theta/(1-\beta+\beta\cos\theta)$.
For $0\le\beta<1/2$, maximizing over $\theta$ bounds the angle by
$\arcsin(\beta/(1-\beta))$, or about $14.48^\circ$ at $\beta=0.2$.
This bound characterizes the angular modification of each unit sensor
feature. The tables separately report the resulting fused predictions.

We use the polar decomposition $B=OH$, with $O$ orthogonal and $H$
symmetric positive definite, to remove one factor at a time from the
same fitted map. Both factors remain the identity outside the observed
span. Table~\ref{tab:ar-factor-diagnostic} shows that each factor changes
classification. These controls decompose a single fitted map, with each
accuracy evaluated separately. Applying the full refined map with
$\beta=1$ gives the changes in
Table~\ref{tab:ar-stage-diagnostic}, using the fit obtained at $\beta=0.2$.
At $\beta=0.2$, feature realignment
corrects 121 raw errors and introduces one with AnyTouch.

\begin{table}[!ht]
 \centering\small
 \caption{Frozen-factor controls, reported as retrospective accuracy (\%).
 Full $B$ applies both polar factors and normalizes each branch separately.
 Common norm. uses the same $B$ but normalizes only after summing the two
 transformed branches. All maps and explicit fusion weights are fixed.}
 \label{tab:ar-factor-diagnostic}
 \input{generated/realignment_factors}
\end{table}


Let $n_m=\|u^mB\|$. The equal-weight serving logits can be written exactly as
\begin{equation}
 \frac{s_c}{\tau}=\frac{u^vBa_c^\top}{2n_v}
                   +\frac{u^tBa_c^\top}{2n_t}.
 \label{eq:ar-branch-gain}
\end{equation}
Relative to the unnormalized transformed branches, the coefficients vary
with the observation. Normalizing these two coefficients to sum to one
gives $\widehat w_v=n_t/(n_v+n_t)$. Omitting their common positive scale
preserves the argmax, but need not preserve softmax probabilities. This is
an equivalent description of the existing normalization, not an additional
reliability estimate. The unit branches are still combined with explicit
weights of one half.

\begin{table}[!ht]
 \centering\small
 \caption{Coefficients relative to unnormalized branches and effects of
 separate branch normalization on the same 704 observations.
 Corrected and Introduced count errors corrected and introduced relative to the
 common-normalization control in Table~\ref{tab:ar-factor-diagnostic}.}
 \label{tab:ar-normalization-diagnostic}
 \input{generated/realignment_normalization}
\end{table}

Separate normalization changes five predictions with AnyTouch, introducing
five errors relative to common normalization
(Table~\ref{tab:ar-normalization-diagnostic}). These retrospective
comparisons quantify how normalization changes the recorded predictions.
Anchor Realignment thus also affects relative branch contributions,
even when the explicit Trust weights are held fixed.

%% file: generated/realignment_values.tex
\newcommand{\arAnyEnergy}{87.66\%}
\newcommand{\arAnyCommonGain}{91.4\%}

%% file: generated/realignment_stages.tex
\begin{tabular}{llrrrr}
\toprule
Backbone & Stage & Correct & Rival & Gap & Accuracy (\%) \\
\midrule
AnyTouch & Raw & 0.1044 & 0.0946 & 0.0098 & 72.30 \\
 & Procrustes, $\beta=.2$ & 0.2843 & 0.2610 & 0.0233 & 81.25 \\
 & CE, $\beta=.2$ & 0.1148 & 0.0823 & 0.0325 & 89.35 \\
 & Same CE map, $\beta=1$ & -0.0001 & -0.0628 & 0.0627 & 94.32 \\
\bottomrule
\end{tabular}

%% file: generated/realignment_factors.tex
\begin{tabular}{lrrrrr}
\toprule
Backbone & Raw & $O$ only & $H$ only & Full $B$ & Common norm. \\
\midrule
AnyTouch & 72.30 & 86.93 & 77.27 & 89.35 & 90.06 \\
\bottomrule
\end{tabular}

%% file: generated/realignment_normalization.tex
\begin{tabular}{lrrr}
\toprule
Backbone & Mean $\widehat w_v$ & Corrected & Introduced \\
\midrule
AnyTouch & 51.46\% & 0 & 5 \\
\bottomrule
\end{tabular}
